\documentclass{article}
\usepackage{style/iclr2027_conference,times}
\usepackage{xspace}

\usepackage{amsmath,amsfonts,bm}

\def\eqref#1{equation~\ref{#1}}

\def\1{\bm{1}}

\DeclareMathAlphabet{\mathsfit}{\encodingdefault}{\sfdefault}{m}{sl}
\SetMathAlphabet{\mathsfit}{bold}{\encodingdefault}{\sfdefault}{bx}{n}

\newcommand{\method}{\textsc{SkillEvoReg}\xspace}
\newcommand{\ccv}{\textsc{CCV}\xspace}

\usepackage{amsmath}
\usepackage{amssymb}
\usepackage{booktabs}
\usepackage{graphicx}
\usepackage{float}
\usepackage{xcolor}
\usepackage{hyperref}
\usepackage{url}
\usepackage{subcaption}
\usepackage{multirow}
\usepackage[table]{xcolor}
\usepackage{array}
\usepackage{algorithm}
\usepackage{algpseudocode}
\usepackage{capt-of}

\algrenewcommand\algorithmicrequire{\textbf{Input:}}
\algrenewcommand\algorithmicensure{\textbf{Output:}}

\definecolor{methodshade}{gray}{0.90}
\newcommand{\ourscell}[1]{\cellcolor{methodshade}#1}
\definecolor{nativeorange}{RGB}{221,143,55}
\definecolor{methodteal}{RGB}{102,196,188}
\definecolor{statebrown}{RGB}{89,19,11}

\usepackage{tikz}
\usetikzlibrary{
    arrows.meta,
    positioning,
    calc,
    fit,
    backgrounds,
    shapes.multipart,
    shapes.misc
}

\newif\ifshowpapertodos
\showpapertodosfalse
\definecolor{papertodoframe}{RGB}{196,105,0}
\definecolor{papertodobg}{RGB}{255,244,224}

\title{SkillEvoReg: Regularizing Agent Skill Evolution Against Overfitting}

\author{
\makebox[\textwidth][c]{%
\begin{tabular}{c}
Guanyu Nie$^{1}$ \quad
Fangzhou Zhu$^{1}$ \quad
Shixiong Kai$^{1}$
\\
Xiongwei Han$^{1}$ \quad
Tao Zhong$^{1}$ \quad
Mingxuan Yuan$^{1}$
\\[4pt]
$^{1}$Huawei Noah's Ark Lab
\\[2pt]
\texttt{nieguanyu@huawei.com}
\end{tabular}%
}
}

\iclrfinalcopy
\begin{document}

\maketitle
\lhead{}
\vspace{-2pt}
\begin{abstract}
Language-model agents increasingly improve by converting execution experience into reusable external skills. Yet repeated skill updates form a learning process of their own: locally useful edits can accumulate into redundant or task-specific instructions, while new updates can disrupt behavior that previously worked. We study this problem as \emph{skill-evolution overfitting} and introduce \method, a general regularization framework for skill evolution inspired by anti-overfitting techniques in neural-network training. \method combines training-time skill dropout, which perturbs update generation, and complexity-aware local regularization, which controls unnecessary structural growth, with causal counterexample validation (CCV), which provides targeted behavioral validation of candidate-specific regressions. We instantiate the framework across heterogeneous skill-evolution systems while retaining each system's native skill evolver and task evaluator. Across SkillOpt, SkillEvolBench, and ContinualSkillBench, \method consistently controls skill-state growth while preserving competitive downstream capability, improves several transfer and later-stage evolution outcomes, and identifies update-level regressions that structural metrics alone cannot reveal. These results suggest that explicit regularization is a useful complement to increasingly capable skill updaters.
\end{abstract}

\section{Introduction}

Large language models (LLMs) are increasingly deployed as agents that interact with environments, invoke tools, and improve from experience through reusable external state \citep{shinn2023reflexion,zhao2024expel,wang2024voyager,zhang2024agentpro,wang2025awm,xu2025amem}. In particular, \emph{agent skills}---reusable instructions, procedures, code, or structured resources that guide future executions---have emerged as a practical substrate for continual self-improvement. Recent systems therefore move beyond one-shot skill generation toward repeated \emph{skill evolution}, treating external skill state as an object that can be revised from execution experience \citep{yang2026skillopt,ni2026trace2skill,zhang2026coevoskills,lei2026skillevolbench,guan2026continualskillbench}.

Repeated skill optimization, however, creates a generalization problem of its own. Each update is inferred from finite and often narrow recent experience, yet the resulting content persists and influences future behavior. Over time, this can produce \emph{structural accumulation} of redundant or narrow rules, \emph{semantic specialization} to incidental properties of recent tasks, or \emph{update-induced behavioral regression} that disrupts behavior already supported by the previous skill. We refer to these phenomena collectively as \emph{skill-evolution overfitting}. Importantly, they need not coincide: substantial structural growth can precede measurable performance loss, while a concise update can introduce a narrow regression without materially increasing skill size.

Recent work has begun to expose or mitigate individual aspects of this problem through consolidation, replay-based validation, and adaptive update control \citep{ni2026trace2skill,guan2026continualskillbench,yang2026gse,he2026skillcommit,li2026skilladam}. We ask a broader question: \emph{can general anti-overfitting principles be instantiated across otherwise heterogeneous skill-evolution procedures?}

This question has a close analogue in neural-network training, where optimization is complemented by mechanisms that perturb learning, control capacity, and test generalization beyond the examples directly optimized. Dropout reduces co-adaptation \citep{srivastava2014dropout}, complexity penalties discourage unnecessary capacity \citep{krogh1991weightdecay}, and augmentation or adversarial training exposes models to informative input variations \citep{goodfellow2015adversarial,zhang2018mixup}. Agent skills are discrete, language-generated artifacts rather than differentiable parameter vectors, so these techniques cannot be transferred mechanically. Their underlying regularization principles, however, provide a useful lens for controlling repeated skill updates.

We introduce \method, a general regularization framework for skill evolution. Rather than prescribing a new updater, \method regularizes both how candidate updates are generated and what structure becomes persistent. \emph{Training-time skill dropout} perturbs the skill context used to generate updates, while \emph{complexity-aware local regularization} controls unnecessary structural growth around candidate changes. Complementing these trajectory-shaping mechanisms, \emph{causal counterexample validation} (CCV) uses candidate-conditioned paired validation to identify update-induced behavioral regressions that structural signals alone cannot reveal. The framework retains each system's native skill evolver and evaluator while adapting these regularization principles to its update semantics.

We evaluate \method across SkillOpt, SkillEvolBench, and ContinualSkillBench, covering single-document optimization, later-stage multi-skill evolution, and continual library growth. Across these settings, \method produces more compact skill states while preserving competitive downstream capability, improves several transfer and later-stage outcomes, and identifies behavioral regressions that structural metrics alone cannot reveal. Component analyses further support the complementary roles of generation-time perturbation, structural regularization, and behavioral validation.

Our contributions are:
\begin{itemize}
\item We introduce \method, a general regularization framework for repeated skill evolution that targets overfitting in the evolution process while retaining the native updater. \method combines training-time skill dropout and complexity-aware local regularization with CCV, which provides targeted behavioral validation for candidate-specific regressions.
\item We instantiate \method across SkillOpt, SkillEvolBench, and ContinualSkillBench, showing that the same regularization principles can be adapted to heterogeneous update semantics while controlling skill-state growth, preserving competitive downstream capability, and improving several transfer and later-stage evolution outcomes.
\end{itemize}
\section{Related Work}

\paragraph{External agent state and skill evolution.} Language-model agents increasingly improve through reusable external state, including reflections, memories, workflows, and executable skills \citep{shinn2023reflexion,zhao2024expel,wang2024voyager,wang2025awm,xu2025amem}. Recent systems go further by treating procedural skill state as an object of iterative optimization or continual maintenance \citep{yang2026skillopt,ni2026trace2skill,zhang2026coevoskills,lei2026skillevolbench,guan2026continualskillbench}. Related prompt- and program-optimization methods similarly search over external language-model state \citep{pryzant2023protegi,yang2024opro,guo2024evoprompt,fernando2024promptbreeder,wang2024promptagent,khattab2024dspy,yuksekgonul2025textgrad}. These lines of work primarily study how to acquire or improve external state; we study how repeated revision of that state should be regularized.

\paragraph{Reliable skill evolution.} Several concurrent methods introduce safeguards against brittle skill updates. GSE combines cross-task consolidation with replay verification \citep{yang2026gse}; SkillCommit validates broader abstractions before commitment \citep{he2026skillcommit}; SkillAdam uses optimization history and adaptive edit budgets \citep{li2026skilladam}; and Trace2Skill consolidates trajectory-local lessons before skill formation \citep{ni2026trace2skill}. These approaches improve reliability within particular evolution procedures. \method instead factors anti-overfitting into a complementary layer that can be instantiated around heterogeneous native updaters, targeting structural accumulation, specialization, and update-induced regression as distinct failure modes.

\paragraph{Regularization and behavioral validation.} Our design draws on the complementary roles of dropout, capacity regularization, and robust validation in conventional learning \citep{srivastava2014dropout,krogh1991weightdecay,goodfellow2015adversarial,zhang2018mixup}. CCV is also related to counterexample-guided repair, where targeted failures expose weaknesses in candidate behavior \citep{orvalho2025counterexample}. Rather than reproducing these techniques mechanically, \method adapts their underlying roles to discrete, language-generated skill transitions. Appendix~\ref{app:extended-related-work} discusses these connections in greater detail.

\section{Skill Evolution as a Regularized Learning Problem}
\label{sec:regularized-skill-evolution}

We consider a generic skill-evolution system with skill state $S_t$, ranging from a single instruction document to a skill library or structured directory. Given an execution trajectory $\tau_t$, the native updater $U$ proposes
\begin{equation}
    \widetilde{S}_{t+1}=U(S_t,\tau_t).
    \label{eq:native-skill-update}
\end{equation}
The host framework then applies its own validation and commit semantics; we assume no particular skill representation, updater, or commit rule.

Repeated application of this update creates a learning problem beyond immediate task performance: each update is inferred from finite recent trajectories yet persists to influence later tasks, allowing locally useful changes to accumulate, specialize, or interfere with previously supported behavior over time.

We use \emph{skill-evolution overfitting} to describe this mismatch between narrow update evidence and persistent downstream influence. Three manifestations are particularly relevant. \emph{Structural accumulation} adds rules, examples, or narrow exceptions without commensurate transferable benefit; \emph{semantic specialization} promotes properties useful for recent trajectories into broadly applied instructions; and \emph{update-induced regression} occurs when a new edit disrupts behavior already supported by the previous skill. These manifestations need not coincide: substantial structure may accumulate before aggregate performance deteriorates, while a concise edit may create a narrow regression without materially increasing skill complexity. Consequently, neither performance degradation nor skill complexity alone is sufficient to characterize skill-evolution overfitting.

\begin{figure*}[t]
    \centering
    \begin{subfigure}[t]{0.32\textwidth}
        \centering
        \includegraphics[width=\linewidth]{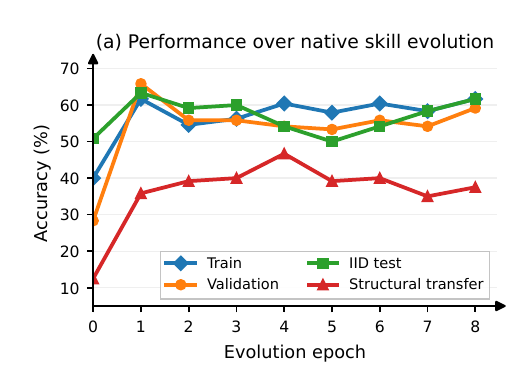}
    \end{subfigure}
    \hfill
    \begin{subfigure}[t]{0.32\textwidth}
        \centering
        \includegraphics[width=\linewidth]{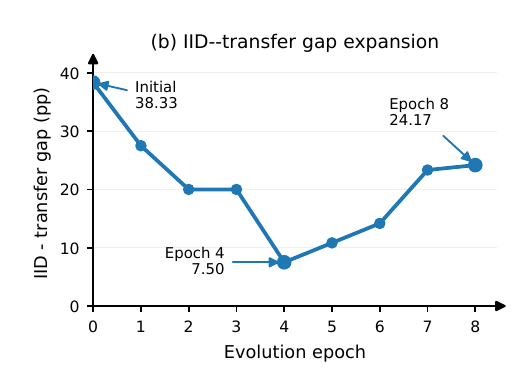}
    \end{subfigure}
    \hfill
    \begin{subfigure}[t]{0.32\textwidth}
        \centering
        \includegraphics[width=\linewidth]{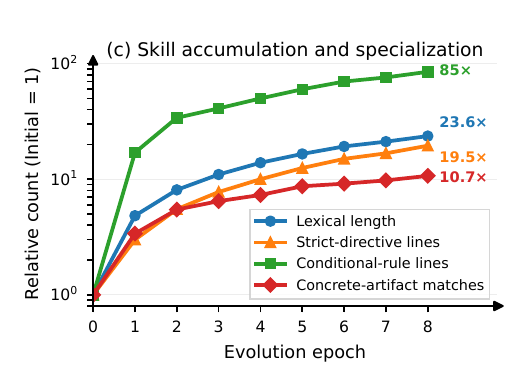}
    \end{subfigure}

    \caption{\textbf{A motivating longitudinal study of native skill evolution on SpreadsheetBench.} \textbf{(a)} Performance across evolution epochs. \textbf{(b)} The IID--transfer gap narrows and then re-expands after the transfer peak. \textbf{(c)} Skill structure continues to accumulate.}
    \label{fig:motivating-longitudinal-study}
\end{figure*}

\subsection{A Motivating Longitudinal Study}
\label{subsec:motivating-longitudinal-study}

We trace a SkillOpt trajectory on SpreadsheetBench for eight epochs, evaluating each checkpoint on an IID test set and a structural transfer-stress set with matched task semantics but greater workbook-level variation. Full split statistics and protocol details are provided in Appendix~\ref{app:motivating-spreadsheet-pilot}.

Figure~\ref{fig:motivating-longitudinal-study} shows a clear post-peak transfer pattern. Transfer performance initially improves and peaks at epoch 4 ($46.67\%$), with the IID--transfer gap shrinking to $7.50$ points. Continued evolution then favors familiar structure: by epoch 8, IID accuracy reaches $61.67\%$ while transfer falls to $37.50\%$, widening the gap to $24.17$ points.

Meanwhile, the skill continues to accumulate more specific structure after transfer has peaked, with lexical length reaching $23.6\times$ its initial value by epoch 8. Later updates thus keep reshaping the skill beyond its best transfer point, motivating explicit regularization of the evolution process.
\newpage
\section{\method}
\label{sec:method}
\label{sec:skillevoreg-method}

\begin{figure*}[t]
    \centering
    \includegraphics[width=\textwidth]{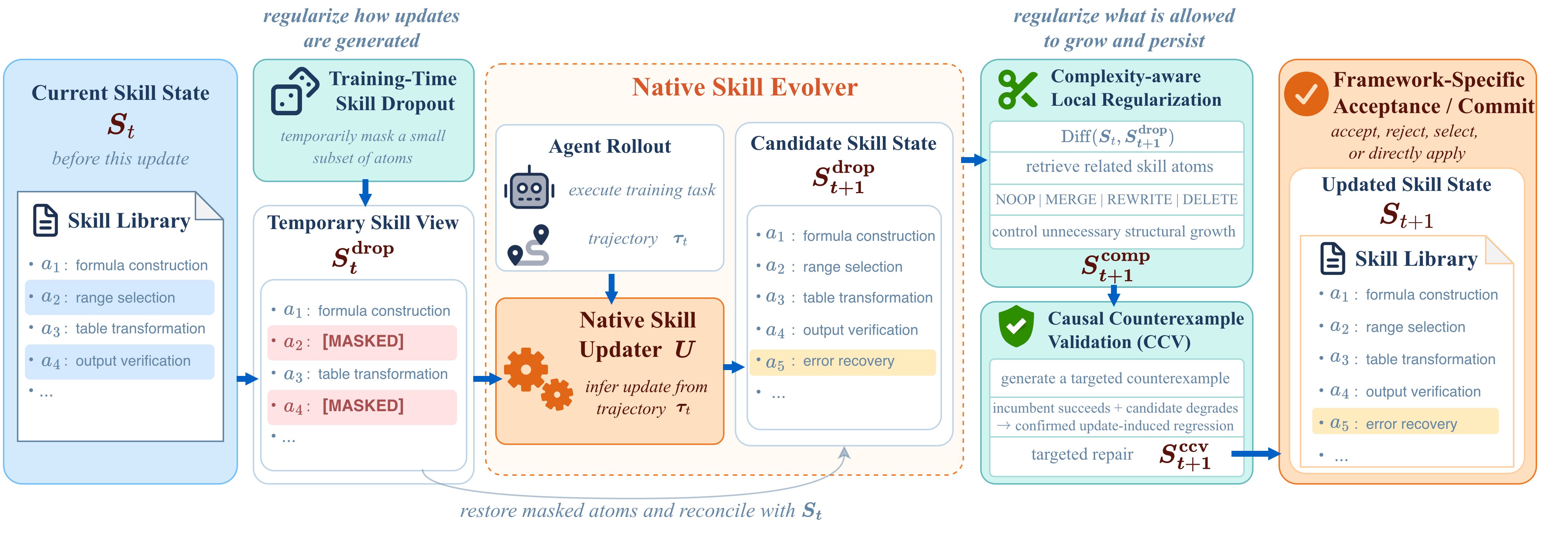}
    \caption{\textbf{Overview of \method.} The three regularizers act on update generation, skill complexity, and behavioral validation, respectively. Colors distinguish the \textcolor{nativeorange}{native skill evolver}, \textcolor{methodteal}{\method regularization modules} and \textcolor{statebrown}{skill-states}.}
    \label{fig:skillevoreg-overview}
\end{figure*}

\subsection{Regularizing Skill Evolution}
\label{subsec:regularizing-skill-evolution}

The pilot suggests that the challenge is not simply producing useful skill updates, but controlling how repeated updates shape the skill over an evolution trajectory. We introduce \method, a general framework that regularizes skill evolution while retaining the native updater. It acts at three complementary points: update generation, skill complexity, and behavioral validation.

The design draws inspiration from anti-overfitting mechanisms in neural-network training: dropout perturbs the information during learning, capacity regularization discourages unnecessary complexity, and validation under informative variations exposes brittle behavior. Because agent skills are discrete language-generated artifacts whose updates persist into future executions, these principles require operational adaptations. \method instantiates the first two roles as training-time skill dropout and complexity-aware local regularization, which directly shape the evolution trajectory, and complements them with CCV, which validates the candidate behavior before persistence.

Let $S_t$ denote the incumbent skill state at update step $t$. We use distinct symbols at regularization-stage boundaries: $S_{t+1}^{\mathrm{drop}}$ is the candidate after dropout reconciliation, $S_{t+1}^{\mathrm{comp}}$ is the candidate after complexity regularization, and $S_{t+1}^{\mathrm{ccv}}$ is the candidate after CCV. The native framework then applies its own acceptance or commit semantics to obtain the next persistent state $S_{t+1}$. Figure~\ref{fig:skillevoreg-overview} shows where the three regularizers intervene. Across evolution systems, the main adaptation lies at the native update boundary, where candidate changes may appear as document edits, library patches, or explicit create/modify operations. Full pseudocode is provided in Appendix~\ref{app:regularization-algorithms}.

\subsection{Training-Time Skill Dropout}
\label{subsec:skillevoreg-dropout}

Skill evolution repeatedly generates new updates while conditioning on the complete current skill state. This can encourage successive updates to depend too strongly on the precise presence or wording of existing instructions. We therefore perturb the skill context used during update generation.

We represent the current skill state as locally meaningful \emph{skill atoms},
\begin{equation}
    \mathcal{A}(S_t)=\{a_1,\ldots,a_m\},
    \label{eq:skill-atoms}
\end{equation}
where an atom is a self-contained instruction, list item, procedural block, or another semantically coherent unit. At update step $t$, we sample $M_t\subseteq\mathcal{A}(S_t)$ and construct a temporary dropout view
\begin{equation}
    S_t^{\mathrm{drop}}
    =
    \operatorname{Mask}(S_t,M_t).
    \label{eq:skill-dropout-view}
\end{equation}
The training rollout and native update are generated under $S_t^{\mathrm{drop}}$. If the native updater proposes $\widetilde{S}_{t+1}^{\mathrm{drop}}$, we extract only the change learned under that perturbed context and reconcile it with the complete incumbent:
\begin{equation}
    \Delta_t^{\mathrm{drop}}
    =
    \operatorname{Diff}
    \left(
        S_t^{\mathrm{drop}},
        \widetilde{S}_{t+1}^{\mathrm{drop}}
    \right),
    \qquad
    S_{t+1}^{\mathrm{drop}}
    =
    \operatorname{Reconcile}
    \left(
        S_t,
        \Delta_t^{\mathrm{drop}}
    \right).
    \label{eq:dropout-reconciliation}
\end{equation}
Thus, masking changes only the context in which an update is learned; masked atoms are restored before later regularization and are never treated as candidate deletions. Frozen evaluation always uses the complete skill state. Algorithm~\ref{alg:skill-dropout} in Appendix~\ref{app:regularization-algorithms} gives the complete procedure.
\paragraph{Remark (Relation to neural-network dropout).}
The analogy lies in when the perturbation acts. Neural dropout perturbs the representation through which a parameter update is learned; skill dropout perturbs the external skill context through which a skill update is generated. In both cases, the perturbation affects learning but is absent from the persistent state used at inference. Unlike inverted neural dropout, no activation-rescaling analogue is needed because skill atoms are discrete contextual units rather than additive activations.

\subsection{Complexity-Aware Local Regularization}
\label{subsec:skillevoreg-complexity}

Skill dropout changes how an update is generated but does not directly control how much structure becomes persistent. Skill state has no differentiable parameter norm, and additional content is not inherently harmful because genuinely new capability may require additional structure. We therefore use observable complexity to determine \emph{when structural reconsideration is needed}, while separately constraining \emph{what content may be changed}, rather than minimizing skill size itself.

Given the incumbent $S_t$ and reconciled candidate $S_{t+1}^{\mathrm{drop}}$, we define the change examined by this stage as
\begin{equation}
    \Delta_t^{\mathrm{comp}}
    =
    \operatorname{Diff}
    \left(
        S_t,
        S_{t+1}^{\mathrm{drop}}
    \right).
    \label{eq:complexity-update-delta}
\end{equation}
Let $\phi(Z)$ summarize observable structural properties of a skill state or update, such as lexical size, atom organization, structural fragmentation, and executable-code structure. The regularizer computes
\begin{equation}
    c_t
    =
    f_{\mathrm{comp}}
    \left(
        \phi(S_t),
        \phi(S_{t+1}^{\mathrm{drop}}),
        \phi(\Delta_t^{\mathrm{comp}})
    \right),
    \label{eq:complexity-signal}
\end{equation}
where $f_{\mathrm{comp}}$ maps these measurements to the framework-specific signal used to trigger or score regularization. Exact feature definitions and realizations of $f_{\mathrm{comp}}$ are given in Appendix~\ref{app:complexity-definitions}.

When structural reconsideration is triggered, we restrict editing to atoms introduced or modified by the candidate and a bounded set of related incumbent atoms:
\begin{equation}
    \mathcal{D}_t
    =
    \operatorname{SegmentAtoms}(\Delta_t^{\mathrm{comp}}),
    \qquad
    \Omega_t
    =
    \mathcal{D}_t
    \cup
    \operatorname{Related}(S_t,\mathcal{D}_t).
    \label{eq:local-regularization-scope}
\end{equation}
This locality constraint prevents a structurally large candidate from triggering an unrestricted rewrite of the complete skill state. A diagnostic SpreadsheetBench case shows that prompt-level locality alone can still yield broad, destructive rewrites, motivating the explicit candidate-local edit scope used here; Appendix~\ref{app:whole-skill-compression-case} provides the detailed analysis.

Inspired by operation-based memory reconciliation in Mem0~\citep{chhikara2025mem0}, we adapt discrete edit decisions to post-update skill reconciliation. For each candidate atom $a\in\mathcal{D}_t$, the local editor selects
\begin{equation}
    o_a
    \in
    \{
    \textsc{NoOp},
    \textsc{Merge},
    \textsc{Rewrite},
    \textsc{Delete}
    \}.
    \label{eq:local-operation-set}
\end{equation}
\textsc{NoOp} preserves useful new structure, while the other operations consolidate, revise, or remove candidate content within $\Omega_t$. Similarity is used only to define what should be considered jointly, not as evidence that two atoms should be merged. We denote the resulting candidate by $S_{t+1}^{\mathrm{comp}}$; if no intervention is triggered, $S_{t+1}^{\mathrm{comp}}=S_{t+1}^{\mathrm{drop}}$. Algorithm~\ref{alg:complexity-regularization} gives the complete procedure.

\paragraph{Remark (Relation to capacity regularization).}
The analogy to neural-network capacity regularization is functional rather than literal. Parametric penalties discourage unnecessary capacity through continuous norms; \method instead uses observable structural complexity to trigger local reconsideration of a discrete candidate update. In both cases the goal is to limit unnecessary capacity accumulated from finite experience, while \textsc{NoOp} explicitly allows additional skill structure when it carries distinct useful behavior. Complexity reduction is therefore a means of regularization, not the objective itself.

\subsection{Causal Counterexample Validation}
\label{subsec:skillevoreg-ccv}

Structural regularization cannot detect every harmful update. A concise candidate may add little complexity while introducing an assumption or decision boundary that breaks behavior already supported by the incumbent. CCV therefore asks whether the structurally regularized candidate $S_{t+1}^{\mathrm{comp}}$ deteriorates on a targeted variation that the incumbent $S_t$ can still handle.

The key comparison is
\begin{equation}
\begin{array}{c|cc}
 & \text{original source } e_t & \text{attacked case } x_t^{\mathrm{ccv}}\\
\hline
\text{incumbent } S_t & \checkmark & \checkmark\\
\text{candidate } S_{t+1}^{\mathrm{comp}} & \checkmark & \times
\end{array}
\quad
\Longrightarrow
\quad
\text{CCV-confirmed regression}.
\label{eq:ccv-four-way-logic}
\end{equation}
Here $\checkmark$ and $\times$ denote evaluator-dependent qualification and degradation rather than necessarily binary success and failure.

CCV constructs this comparison in three steps. First, it selects an eligible source from the available training history $\mathcal{H}_t$,
\begin{equation}
    e_t
    =
    \operatorname{SelectSource}
    \left(
        \mathcal{H}_t,
        S_t,
        S_{t+1}^{\mathrm{comp}}
    \right).
    \label{eq:ccv-source-selection}
\end{equation}
The incumbent must already support $e_t$, and the candidate must remain qualified on the same unmodified source. This clean-pair requirement rules out candidates that have already broken the original task.

Second, because CCV acts after complexity regularization, it defines its own stage-local candidate change and uses it to construct the attack:
\begin{equation}
\Delta_t^{\mathrm{ccv}}=\operatorname{Diff}(S_t,S_{t+1}^{\mathrm{comp}}),\quad a_t=\operatorname{GenerateAttack}(e_t,\Delta_t^{\mathrm{ccv}}),\quad x_t^{\mathrm{ccv}}=\operatorname{ApplyAttack}(e_t,a_t).
\label{eq:ccv-attack-generation}
\end{equation}
Here $a_t$ specifies a targeted perturbation of an assumption, dependency, or decision boundary exposed by $\Delta_t^{\mathrm{ccv}}$, and $x_t^{\mathrm{ccv}}$ is the resulting attacked case.

Third, the incumbent and candidate are evaluated independently on that same case:
\begin{equation}
\begin{aligned}
r_{\mathrm{old}}=&\operatorname{Evaluate}(S_t,x_t^{\mathrm{ccv}}),\qquad
r_{\mathrm{new}}=\operatorname{Evaluate}(S_{t+1}^{\mathrm{comp}},x_t^{\mathrm{ccv}}),\\
&g_t=\operatorname{RegressionCriterion}(r_{\mathrm{old}},r_{\mathrm{new}};\theta_{\mathrm{ccv}}).
\end{aligned}
\label{eq:ccv-paired-evaluation}
\end{equation}
where $\theta_{\mathrm{ccv}}$ denotes evaluator-specific qualification and degradation criteria. A regression is confirmed only when the incumbent remains sufficiently successful on the attacked case while the candidate deteriorates. When $g_t=1$, CCV performs at most one scope-constrained repair over the implicated candidate region; otherwise the candidate is unchanged. We denote the resulting CCV-processed candidate by $S_{t+1}^{\mathrm{ccv}}$. Exact qualification rules, attack construction, and repair procedures are given in Appendix~\ref{app:ccv-algorithm}.

\paragraph{Remark (Relation to data augmentation).}
CCV adapts the augmentation principle to a skill transition: the candidate update itself reveals where generalization may have narrowed, and the attack perturbs the source along that direction. The resulting case is therefore a targeted stress test of the behavioral boundary moved by the update, used primarily for paired validation and, when needed, scope-constrained repair rather than simply being added back to training.

\paragraph{Remark (Why causal?).}
The four-way comparison in Equation~\ref{eq:ccv-four-way-logic} provides the basis for the causal terminology. Qualification of both skills on the original source controls for ordinary source-task failure, while incumbent success on the attacked case controls for an attack that is simply too difficult. Candidate-specific degradation on that same case therefore associates the regression with the candidate transition. We use \emph{CCV-confirmed regression} in this operational sense, not as a claim of formal causal identification under stochastic rollouts or multi-edit candidates.

\subsection{Adapting to Different Update Semantics}
\label{subsec:update-semantics}

The regularizers retain the same roles across evolution systems; the main adaptation lies at the native update boundary. SkillOpt exposes ranked edits to a single skill document, SkillEvolBench localized patches over a multi-skill library, and ContinualSkillBench explicit \texttt{CREATE}/\texttt{MODIFY} operations. These semantics determine how stage-specific candidate changes are exposed, how dropout updates are reconciled with the incumbent, how editable scopes are defined, and how $S_{t+1}^{\mathrm{ccv}}$ is returned to the native framework. Appendix~\ref{app:regularization-algorithms} details the framework-specific integrations.
\section{Experiments}
\label{sec:experiments}

We evaluate whether the same regularization principles improve three distinct forms of skill evolution: repeated optimization of a single skill, later-stage revision of an established multi-skill library, and native continual skill-library growth. We then analyze the roles of the individual regularization components.

\subsection{Experimental Setup}
\label{sec:exp-setup}

All compared methods use the same proprietary instruction-following LLM under a fixed inference configuration and each benchmark's native execution harness, task order, and evaluator. \method is inserted at the candidate-update boundary while retaining the native skill evolver and evaluation protocol. We report downstream capability together with skill-state size and structural complexity $C$; where checkpoint selection is available, we distinguish the validation-selected \emph{delivery} checkpoint from the terminal \emph{final} checkpoint. Full benchmark protocols and measurement conventions are given in Appendix~\ref{app:skillopt-measurement}--\ref{app:evaluation-conventions}. Frozen-evaluation inference usage is analyzed separately in Appendix~\ref{app:inference-cost} as a deployment-side by-product of compact persistent skill states rather than an optimization objective of \method.

\subsection{SkillOpt: Episodic Single-Skill Optimization}
\label{sec:exp-skillopt}

SkillOpt~\citep{yang2026skillopt} repeatedly optimizes a single instruction document. We evaluate SpreadsheetBench, SearchQA, and LiveMathBench over eight evolution epochs, retaining SkillOpt's native updater and checkpointing procedure while regularizing candidate transitions. Full protocol and checkpoint details are given in Appendix~\ref{app:skillopt-measurement}.

\begin{table}[ht]
\caption{SkillOpt results. Delivery denotes the validation-selected checkpoint and Final the epoch-8 checkpoint; $\Delta$ Test is Final minus Delivery. Skill tokens and complexity describe the final skill state.}
\label{tab:exp-skillopt}
\centering
\footnotesize
\setlength{\tabcolsep}{3.5pt}
\begin{tabular}{clrrrrr}
\toprule
\textbf{Dataset} & \textbf{Method}
& \textbf{\shortstack{Delivery\\Test $\uparrow$}}
& \textbf{\shortstack{Final\\Test $\uparrow$}}
& \textbf{$\Delta$ Test $\uparrow$}
& \textbf{\shortstack{Skill\\tokens $\downarrow$}}
& \textbf{Complexity $\downarrow$} \\
\midrule
\multirow[c]{2}{*}{Spreadsheet}
& SkillOpt & 54.75 & 47.00 & $-7.75$ & 6,060 & 116.96 \\
& \ourscell{\method}
& \ourscell{\textbf{58.50}}
& \ourscell{\textbf{59.25}}
& \ourscell{$\mathbf{+0.75}$}
& \ourscell{\textbf{2,341}}
& \ourscell{\textbf{57.21}} \\
\midrule
\multirow[c]{2}{*}{SearchQA}
& SkillOpt & 80.46 & \textbf{80.86} & $\mathbf{+0.39}$ & 7,924 & 124.42 \\
& \ourscell{\method}
& \ourscell{\textbf{81.14}}
& \ourscell{80.39}
& \ourscell{$-0.75$}
& \ourscell{\textbf{1,833}}
& \ourscell{\textbf{39.51}} \\
\midrule
\multirow[c]{2}{*}{LiveMath}
& SkillOpt & 34.79 & \textbf{36.41} & $\mathbf{+1.61}$ & 6,749 & 110.52 \\
& \ourscell{\method}
& \ourscell{\textbf{35.02}}
& \ourscell{30.18}
& \ourscell{$-4.84$}
& \ourscell{\textbf{1,606}}
& \ourscell{\textbf{36.52}} \\
\bottomrule
\end{tabular}
\end{table}

\paragraph{Capability and evolution quality.} At the delivery checkpoint, \method matches or improves all three tasks while producing substantially smaller final skill states. SpreadsheetBench continues to improve through the terminal checkpoint, while SearchQA remains essentially stable. LiveMath instead exposes a distribution-sensitive trade-off. SkillOpt repeatedly reinforces a meta-option prior favoring the answer that ``a stronger result can be proved,'' whereas \method consolidates repeated reinforcement of this pattern. Of the 217 test questions, 92 use this meta-option as the correct answer, and this subset accounts for $85.7\%$ of \method's late-stage decline. The performance gap is therefore strongly concentrated in a specific test-distribution pattern rather than distributed uniformly across questions. Appendix~\ref{app:livemath-shortcut} provides the item-level analysis.

\subsection{SkillEvolBench: Regularizing Later-Stage Skill Evolution}
\label{sec:exp-skillevol}

SkillEvolBench~\citep{lei2026skillevolbench} evaluates transfer under context shift, adversarial variation, and skill composition. Its native protocol uses a single acquisition pass. We add a controlled second pass to emulate later-stage skill evolution, where an already formed library continues to absorb repeated experience and may begin to over-specialize. Starting from the same native one-pass library, we compare freezing it (1-pass), continuing with a second unregularized pass (2-pass), and regularizing that same second pass with \method. The two second-pass conditions therefore receive identical additional experience, while deployment tasks remain unseen throughout acquisition. Details are given in Appendix~\ref{app:framework-instantiations} and Appendix~\ref{app:skillevol-measurement}.

\noindent
\begin{minipage}[t]{0.43\linewidth}
\vspace{0pt}
\paragraph{Later-stage transfer and controlled growth.} A second unregularized pass illustrates the risk of repeated skill evolution: the library continues to grow, while deployment success decreases from $31.67\%$ to $29.44\%$. Applying \method to the same additional experience instead reaches $33.33\%$ deployment success while substantially controlling persistent-state growth. Relative to the one-pass anchor, \method reduces second-pass token growth by $34.0\%$ and complexity growth by $32.7\%$ compared with unregularized two-pass evolution.
\end{minipage}
\hfill
\begin{minipage}[t]{0.53\linewidth}
\centering
\captionsetup{hypcap=false}
\captionof{table}{SkillEvolBench results. T1--T3 are replay tasks and T4--T6 are unseen deployment tasks.}
\label{tab:exp-skillevol}
\vspace{0pt}
\small
\setlength{\tabcolsep}{3.2pt}
\renewcommand{\arraystretch}{1.02}
\begin{tabular}{@{}lrr>{\columncolor{methodshade}}r@{}}
\toprule
\textbf{Metric} & \textbf{1-pass} & \textbf{2-pass} & \textbf{Ours} \\
\midrule
Replay (T1--T3) $\uparrow$ & 40.00 & 38.89 & \textbf{42.22} \\
\quad T1 Canonical $\uparrow$ & 43.33 & 40.00 & \textbf{48.33} \\
\quad T2 Enriched $\uparrow$ & \textbf{41.67} & 36.67 & 38.33 \\
\quad T3 Variant $\uparrow$ & 35.00 & \textbf{40.00} & \textbf{40.00} \\
\midrule
Deployment (T4--T6) $\uparrow$ & 31.67 & 29.44 & \textbf{33.33} \\
\quad T4 Context shift $\uparrow$ & 36.67 & \textbf{40.00} & 31.67 \\
\quad T5 Adversarial $\uparrow$ & 40.00 & 35.00 & \textbf{41.67} \\
\quad T6 Composition $\uparrow$ & 18.33 & 13.33 & \textbf{26.67} \\
\midrule
Skill tokens $\downarrow$ & 30,828 & 36,921 & 34,847 \\
Complexity $C$ $\downarrow$ & 609.20 & 696.46 & 667.97 \\
\bottomrule
\end{tabular}
\end{minipage}

\paragraph{Transfer behavior.} The improvement is not uniform across every transfer mode. \method gives the strongest gain on composition ($26.67\%$ versus $18.33\%$ after one pass and $13.33\%$ after two unregularized passes) and improves adversarial transfer, while context-shift success decreases. Together with the controlled skill-state growth, this suggests that regularization does not simply suppress second-pass learning, but changes which parts of repeated experience are allowed to persist. Continuous verifier scores and additional structural diagnostics are reported in Appendix~\ref{app:skillevol-measurement}.

\subsection{ContinualSkillBench: Continual Skill-Library Evolution}
\label{sec:exp-continual}

ContinualSkillBench~\citep{guan2026continualskillbench} extends the setting further by allowing the evolving library to create new skills as well as modify existing ones through native \textsc{Create-Skill} and \textsc{Modify-Skill} actions. This makes library proliferation itself part of the regularization problem. Each condition learns sequentially on tasks 1--50, freezes the resulting library, and evaluates tasks 51--100 without further learning. \method regularizes both CREATE and MODIFY outcomes before write-back while retaining the benchmark's native update semantics. Complete protocol and online-learning results are given in Appendix~\ref{app:continual-measurement} and Appendix~\ref{app:framework-instantiations}.

\begin{table}[ht]
\caption{ContinualSkillBench frozen-library results after learning tasks 1--50. Scores average two evaluations of held-out tasks 51--100; skill-state statistics use the complete step-50 library.}
\label{tab:exp-continual}
\centering
\small
\setlength{\tabcolsep}{3.0pt}
\renewcommand{\arraystretch}{1.04}
\begin{tabular}{@{}ccrrr@{\hspace{7pt}}rrr@{}}
\toprule
\multirow[c]{3}{*}{\textbf{Domain}} & \multirow[c]{3}{*}{\textbf{Method}} & \multicolumn{3}{c}{\textbf{Held-out capability $\uparrow$}} & \multicolumn{3}{c}{\textbf{Skill state $\downarrow$}} \\
\cmidrule(lr){3-5}\cmidrule(l){6-8}
& & \textbf{Overall} & \textbf{Struct.} & \textbf{Rubric} & \textbf{\# Skills} & \textbf{Tokens} & \textbf{Complexity} \\
\midrule
\multirow[c]{2}{*}{Mathematics} & Baseline & 67.90 & 74.00 & 61.80 & 33 & 21,952 & 612.26 \\
& \ourscell{\method} & \ourscell{\textbf{69.50}} & \ourscell{74.00} & \ourscell{\textbf{65.00}} & \ourscell{\textbf{32}} & \ourscell{\textbf{16,368}} & \ourscell{\textbf{438.30}} \\
\midrule
\multirow[c]{2}{*}{Law} & Baseline & 65.42 & 81.82 & 60.79 & 24 & 25,826 & 645.78 \\
& \ourscell{\method} & \ourscell{\textbf{70.69}} & \ourscell{\textbf{86.36}} & \ourscell{\textbf{66.27}} & \ourscell{\textbf{20}} & \ourscell{\textbf{19,941}} & \ourscell{\textbf{507.97}} \\
\midrule
\multirow[c]{2}{*}{Finance} & Baseline & 68.63 & 65.58 & 69.59 & 28 & 78,193 & 1,596.72 \\
& \ourscell{\method} & \ourscell{\textbf{69.10}} & \ourscell{\textbf{65.95}} & \ourscell{\textbf{70.09}} & \ourscell{\textbf{17}} & \ourscell{\textbf{71,715}} & \ourscell{\textbf{1,436.19}} \\
\midrule
\multirow[c]{2}{*}{Office} & Baseline & \textbf{82.49} & \textbf{78.57} & \textbf{85.33} & 18 & 19,038 & 502.45 \\
& \ourscell{\method} & \ourscell{81.05} & \ourscell{76.19} & \ourscell{84.57} & \ourscell{\textbf{17}} & \ourscell{\textbf{16,087}} & \ourscell{\textbf{414.60}} \\
\midrule
\multirow[c]{2}{*}{Healthcare} & Baseline & 70.92 & \textbf{66.67} & 72.27 & 42 & 33,646 & 822.07 \\
& \ourscell{\method} & \ourscell{\textbf{73.12}} & \ourscell{62.50} & \ourscell{\textbf{76.47}} & \ourscell{\textbf{34}} & \ourscell{\textbf{28,416}} & \ourscell{\textbf{727.60}} \\
\bottomrule
\end{tabular}
\end{table}

\paragraph{Held-out capability and consolidation.} \method maintains comparable or better Overall held-out performance across the five domains while consistently producing more compact persistent skill states. Skill tokens, complexity, and skill count decrease in every domain. The reduction is not simply uniform pruning: Mathematics primarily shortens existing skills, whereas the other domains show stronger suppression of library proliferation. Together, these results show that regularization controls persistent-state accumulation without broadly sacrificing held-out capability. Evaluator-level breakdowns and sensitivity analyses are provided in Appendix~\ref{app:continual-measurement}.

\subsection{Component Analysis}
\label{sec:exp-analysis}

We next examine how dropout and complexity regularization shape the evolution trajectory, and how CCV complements them with targeted behavioral validation.

\subsubsection{Dropout and Complexity Regularization Are Complementary}
\label{sec:analysis-ablation}

\noindent
\begin{minipage}[t]{0.57\linewidth}
\vspace{0pt}
To isolate the roles of skill dropout and complexity-aware regularization, we conduct a controlled SpreadsheetBench ablation with four variants: the native updater, complexity regularization only, skill dropout only, and the full \method. All variants use the same eight-epoch evolution budget, training data, native updater, validation procedure, and frozen evaluation; only the active regularizers differ, and we compare final checkpoints to ensure equal evolution budgets.

\end{minipage}
\hfill
\begin{minipage}[t]{0.39\linewidth}
\vspace{0pt}
\centering
\captionsetup{hypcap=false}
\captionof{table}{SpreadsheetBench ablation at the final checkpoint.}
\label{tab:ablation}
\vspace{3pt}
\small
\setlength{\tabcolsep}{4pt}
\begin{tabular}{lrr}
\toprule
\textbf{Method} &
\textbf{Test $\uparrow$} &
\textbf{Chars. $\downarrow$} \\
\midrule
Native updater & 47.00 & 31.7K \\
Complexity only & 55.50 & 13.1K \\
Dropout only & 57.50 & 19.5K \\
\rowcolor{methodshade}
Full \method & \textbf{59.25} & \textbf{11.7K} \\
\bottomrule
\end{tabular}
\end{minipage}

The two regularizers exhibit complementary effects. Complexity regularization substantially limits persistent-state growth, while dropout alone attains higher test accuracy but allows a larger skill state. Their combination yields both the smallest regularized skill and the strongest frozen score, consistent with dropout shaping how updates are generated and complexity regularization controlling what structure is allowed to persist. \ccv{} does not modify the persistent state on this particular trajectory, so we examine its behavioral role separately.

\subsubsection{\ccv{} Detects Regressions Beyond Structural Signals}
\label{sec:analysis-ccv}

\ccv{} provides a complementary behavioral signal: even a concise update can disrupt behavior already supported by the incumbent without producing a distinctive structural signature. Across the evaluated trajectories, we identify 12 distinct CCV regression signals and 9 state-changing repairs. A representative SkillEvolBench dependency-resolution trace illustrates the mechanism: after the incumbent and candidate both qualify on the original source, a candidate-conditioned package-boundary attack yields reward $1.0$ for the incumbent and $0.8$ for the candidate. CCV identifies candidate-specific sensitivity to the package-boundary condition and produces a scope-constrained rewrite; a later replay recovers the affected process checks. Appendix~\ref{app:ccv-details} provides additional regression evidence and a detailed representative trace.

\section{Conclusion}
Repeated skill evolution creates a learning problem beyond the native updater. We introduced \method, a general framework that translates complementary anti-overfitting principles from neural-network training to discrete skill evolution: training-time skill dropout perturbs update generation, complexity-aware local regularization controls persistent structural growth, and CCV provides targeted behavioral validation of candidate-specific regressions. Across SkillOpt, SkillEvolBench, and ContinualSkillBench, \method limits skill-state growth while preserving downstream capability and improving several transfer and later-stage outcomes. More broadly, these results suggest that regularization principles developed for parametric learning provide a useful lens for controlling persistent, language-generated skill updates, and support regularization as a first-class component of skill evolution.

\bibliography{refs}

@inproceedings{chhikara2025mem0,
  title={{Mem0}: Building Production-Ready {AI} Agents with Scalable Long-Term Memory},
  author={Chhikara, Prateek and Khant, Dev and Aryan, Saket and Singh, Taranjeet and Yadav, Deshraj},
  booktitle={Proceedings of the 28th European Conference on Artificial Intelligence},
  year={2025}
}

@inproceedings{shinn2023reflexion,
  title     = {Reflexion: Language Agents with Verbal Reinforcement Learning},
  author    = {Shinn, Noah and Cassano, Federico and Gopinath, Ashwin and
               Narasimhan, Karthik and Yao, Shunyu},
  booktitle = {Advances in Neural Information Processing Systems},
  volume    = {36},
  year      = {2023}
}

@inproceedings{zhao2024expel,
  title     = {{ExpeL}: {LLM} Agents Are Experiential Learners},
  author    = {Zhao, Andrew and Huang, Daniel and Xu, Quentin and Lin, Matthieu and
               Liu, Yong-Jin and Huang, Gao},
  booktitle = {Proceedings of the AAAI Conference on Artificial Intelligence},
  volume    = {38},
  pages     = {19632--19642},
  year      = {2024}
}

@article{wang2024voyager,
  title   = {Voyager: An Open-Ended Embodied Agent with Large Language Models},
  author  = {Wang, Guanzhi and Xie, Yuqi and Jiang, Yunfan and Mandlekar, Ajay and
             Xiao, Chaowei and Zhu, Yuke and Fan, Linxi and Anandkumar, Anima},
  journal = {Transactions on Machine Learning Research},
  year    = {2024}
}

@inproceedings{zhang2024agentpro,
  title     = {Agent-Pro: Learning to Evolve via Policy-Level Reflection and Optimization},
  author    = {Zhang, Wenqi and Tang, Ke and Wu, Hai and Wang, Mengna and
               Shen, Yongliang and Hou, Guiyang and Tan, Zeqi and Li, Peng and
               Zhuang, Yueting and Lu, Weiming},
  booktitle = {Proceedings of the 62nd Annual Meeting of the Association for Computational Linguistics},
  pages     = {5348--5375},
  year      = {2024},
  publisher = {Association for Computational Linguistics}
}

@inproceedings{wang2025awm,
  title     = {Agent Workflow Memory},
  author    = {Wang, Zora Zhiruo and Mao, Jiayuan and Fried, Daniel and Neubig, Graham},
  booktitle = {Proceedings of the 42nd International Conference on Machine Learning},
  series    = {Proceedings of Machine Learning Research},
  volume    = {267},
  pages     = {63897--63911},
  year      = {2025},
  publisher = {PMLR}
}

@inproceedings{xu2025amem,
  title     = {{A-Mem}: Agentic Memory for {LLM} Agents},
  author    = {Xu, Wujiang and Liang, Zujie and Mei, Kai and Gao, Hang and
               Tan, Juntao and Zhang, Yongfeng},
  booktitle = {Advances in Neural Information Processing Systems},
  volume    = {38},
  year      = {2025}
}

@misc{yang2026skillopt,
      title={SkillOpt: Executive Strategy for Self-Evolving Agent Skills}, 
      author={Yifan Yang and Ziyang Gong and Weiquan Huang and Qihao Yang and Ziwei Zhou and Zisu Huang and Yan Li and Xuemei Gao and Qi Dai and Bei Liu and Kai Qiu and Yuqing Yang and Dongdong Chen and Xue Yang and Chong Luo},
      year={2026},
      eprint={2605.23904},
      archivePrefix={arXiv},
      primaryClass={cs.AI},
      url={https://arxiv.org/abs/2605.23904}, 
}

@misc{ni2026trace2skill,
      title={Trace2Skill: Distill Trajectory-Local Lessons into Transferable Agent Skills}, 
      author={Jingwei Ni and Yihao Liu and Xinpeng Liu and Yutao Sun and Mengyu Zhou and Pengyu Cheng and Dexin Wang and Erchao Zhao and Xiaoxi Jiang and Guanjun Jiang},
      year={2026},
      eprint={2603.25158},
      archivePrefix={arXiv},
      primaryClass={cs.AI},
      url={https://arxiv.org/abs/2603.25158}, 
}

@misc{zhang2026coevoskills,
      title={{CoEvoSkills}: Self-Evolving Agent Skills via Co-Evolutionary Verification}, 
      author={Hanrong Zhang and Shicheng Fan and Henry Peng Zou and Yankai Chen and Zhenting Wang and Jiayu Zhou and Chengze Li and Wei-Chieh Huang and Yifei Yao and Kening Zheng and Xue and Liu and Xiaoxiao Li and Philip S. Yu},
      year={2026},
      eprint={2604.01687},
      archivePrefix={arXiv},
      primaryClass={cs.AI},
      url={https://arxiv.org/abs/2604.01687}, 
}

@misc{lei2026skillevolbench,
      title={SkillEvolBench: Benchmarking the Evolution from Episodic Experience to Procedural Skills}, 
      author={Yingtie Lei and Zhongwei Wan and Jiankun Zhang and Samiul Alam and Zixuan Zhong and Peizhou Huang and Xin Wang and Jingxuan Zhang and Donghao Zhou and Yunta Hsieh and Zhihao Dou and Hui Shen and Yan Xu and Dimitrios Dimitriadis and Tuo Zhang and Mi Zhang},
      year={2026},
      eprint={2605.24117},
      archivePrefix={arXiv},
      primaryClass={cs.AI},
      url={https://arxiv.org/abs/2605.24117}, 
}

@misc{guan2026continualskillbench,
      title={ContinualSkillBench: Can LLM Agents Truly Evolve Their Capabilities?}, 
      author={Tianyi Guan and Yiding Wang and Haotong Yang and Siyuan Cao and Shirui Liu and Yi Hu and Jiaqi Li and Muhan Zhang},
      year={2026},
      eprint={2608.03874},
      archivePrefix={arXiv},
      primaryClass={cs.AI},
      url={https://arxiv.org/abs/2608.03874}, 
}

@misc{yang2026gse,
      title={Learning Globally Reusable Skills for Coding Agents}, 
      author={Chen Yang and Jiashuo Tian and Ziqi Wang and Xinyin Liu and Meiru Ye and Junjie Chen},
      year={2026},
      eprint={2608.06153},
      archivePrefix={arXiv},
      primaryClass={cs.SE},
      url={https://arxiv.org/abs/2608.06153}, 
}

@misc{he2026skillcommit,
      title={SkillCommit: Evolving Agent Skills through Behaviorally Validated Scope Expansion}, 
      author={Yu He and Weikai Yang},
      year={2026},
      eprint={2608.15165},
      archivePrefix={arXiv},
      primaryClass={cs.AI},
      url={https://arxiv.org/abs/2608.15165}, 
}

@misc{li2026skilladam,
      title={SkillAdam: Stable and Efficient Skill Evolution for Agents}, 
      author={Gaoyuan Li and Meihao Fan and Yizhe Liu and Shaolei Zhang and Ju Fan and Siyi Wang and Jiaheng Hou and Xudong Weng and Honghan Tian and Zang Li},
      year={2026},
      eprint={2609.08944},
      archivePrefix={arXiv},
      primaryClass={cs.AI},
      url={https://arxiv.org/abs/2609.08944}, 
}

@inproceedings{pryzant2023protegi,
  title     = {Automatic Prompt Optimization with ``Gradient Descent'' and Beam Search},
  author    = {Pryzant, Reid and Iter, Dan and Li, Jerry and Lee, Yin and
               Zhu, Chenguang and Zeng, Michael},
  booktitle = {Proceedings of the 2023 Conference on Empirical Methods in Natural Language Processing},
  pages     = {7957--7968},
  year      = {2023},
  publisher = {Association for Computational Linguistics}
}

@inproceedings{yang2024opro,
  title     = {Large Language Models as Optimizers},
  author    = {Yang, Chengrun and Wang, Xuezhi and Lu, Yifeng and Liu, Hanxiao and
               Le, Quoc V. and Zhou, Denny and Chen, Xinyun},
  booktitle = {International Conference on Learning Representations},
  year      = {2024}
}

@inproceedings{guo2024evoprompt,
  title     = {Connecting Large Language Models with Evolutionary Algorithms Yields Powerful Prompt Optimizers},
  author    = {Guo, Qingyan and Wang, Rui and Guo, Junliang and Li, Bei and
               Song, Kaitao and Tan, Xu and Liu, Guoqing and Bian, Jiang and Yang, Yujia},
  booktitle = {International Conference on Learning Representations},
  year      = {2024}
}

@inproceedings{wang2024promptagent,
  title     = {{PromptAgent}: Strategic Planning with Language Models Enables Expert-Level Prompt Optimization},
  author    = {Wang, Xinyuan and Li, Chenxi and Wang, Zhen and Bai, Fan and
               Luo, Haotian and Zhang, Jiayou and Jojic, Nebojsa and
               Xing, Eric P. and Hu, Zhiting},
  booktitle = {International Conference on Learning Representations},
  year      = {2024}
}

@inproceedings{fernando2024promptbreeder,
  title     = {{Promptbreeder}: Self-Referential Self-Improvement via Prompt Evolution},
  author    = {Fernando, Chrisantha and Banarse, Dylan Sunil and Michalewski, Henryk and
               Osindero, Simon and Rockt{\"a}schel, Tim},
  booktitle = {Proceedings of the 41st International Conference on Machine Learning},
  series    = {Proceedings of Machine Learning Research},
  volume    = {235},
  pages     = {13481--13544},
  year      = {2024},
  publisher = {PMLR}
}

@inproceedings{khattab2024dspy,
  title     = {{DSPy}: Compiling Declarative Language Model Calls into State-of-the-Art Pipelines},
  author    = {Khattab, Omar and Singhvi, Arnav and Maheshwari, Paridhi and
               Zhang, Zhiyuan and Santhanam, Keshav and A, Sri Vardhamanan and
               Haq, Saiful and Sharma, Ashutosh and Joshi, Thomas T. and
               Moazam, Hanna and Miller, Heather and Zaharia, Matei and Potts, Christopher},
  booktitle = {International Conference on Learning Representations},
  year      = {2024}
}

@article{yuksekgonul2025textgrad,
  title   = {Optimizing Generative {AI} by Backpropagating Language Model Feedback},
  author  = {Yuksekgonul, Mert and Bianchi, Federico and Boen, Joseph and
             Liu, Sheng and Lu, Pan and Huang, Zhi and Guestrin, Carlos and Zou, James},
  journal = {Nature},
  volume  = {639},
  pages   = {609--616},
  year    = {2025}
}

@article{srivastava2014dropout,
  title   = {Dropout: A Simple Way to Prevent Neural Networks from Overfitting},
  author  = {Srivastava, Nitish and Hinton, Geoffrey and Krizhevsky, Alex and
             Sutskever, Ilya and Salakhutdinov, Ruslan},
  journal = {Journal of Machine Learning Research},
  volume  = {15},
  number  = {56},
  pages   = {1929--1958},
  year    = {2014}
}

@inproceedings{krogh1991weightdecay,
  title     = {A Simple Weight Decay Can Improve Generalization},
  author    = {Krogh, Anders and Hertz, John A.},
  booktitle = {Advances in Neural Information Processing Systems},
  volume    = {4},
  pages     = {950--957},
  year      = {1991}
}

@inproceedings{goodfellow2015adversarial,
  title     = {Explaining and Harnessing Adversarial Examples},
  author    = {Goodfellow, Ian J. and Shlens, Jonathon and Szegedy, Christian},
  booktitle = {International Conference on Learning Representations},
  year      = {2015}
}

@inproceedings{zhang2018mixup,
  title     = {mixup: Beyond Empirical Risk Minimization},
  author    = {Zhang, Hongyi and Ciss{\'e}, Moustapha and Dauphin, Yann N. and Lopez-Paz, David},
  booktitle = {International Conference on Learning Representations},
  year      = {2018}
}

@incollection{prechelt1998earlystopping,
  title     = {Early Stopping---But When?},
  author    = {Prechelt, Lutz},
  booktitle = {Neural Networks: Tricks of the Trade},
  editor    = {Orr, Genevieve B. and M{\"u}ller, Klaus-Robert},
  series    = {Lecture Notes in Computer Science},
  volume    = {1524},
  pages     = {55--69},
  publisher = {Springer},
  address   = {Berlin, Heidelberg},
  year      = {1998}
}

@inproceedings{orvalho2025counterexample,
  title     = {Counterexample Guided Program Repair Using Zero-Shot Learning and {MaxSAT}-Based Fault Localization},
  author    = {Orvalho, Pedro and Janota, Mikol{\'a}{\v{s}} and Manquinho, Vasco M.},
  booktitle = {Proceedings of the AAAI Conference on Artificial Intelligence},
  volume    = {39},
  pages     = {649--657},
  year      = {2025}
}
\bibliographystyle{style/iclr2027_conference}

\newpage
\appendix
\section{Additional Experimental Details}
\label{app:additional-experimental-details}

\subsection{Motivating SpreadsheetBench Pilot}
\label{app:motivating-spreadsheet-pilot}

\paragraph{Purpose.}
The study in Section~\ref{subsec:motivating-longitudinal-study} is a longitudinal stress test of native skill evolution rather than a confirmatory evaluation of \method. The pilot is designed to examine how continued skill updates affect performance on familiar workbook structures and on structurally shifted workbooks while the skill state itself evolves. We therefore use this study as motivating evidence for the regularization problem; all comparative claims about \method are based on the independently specified experiments in Section~\ref{sec:experiments}.

\paragraph{IID and structural-transfer evaluation.}
We organize the held-out evaluation into an in-distribution (IID) set and a structural transfer-stress set. The two sets keep the primary task semantics closely matched while differing in workbook-level structure. They contain identical proportions of the benchmark's primary operation families and the same proportions of cell-level and sheet-level tasks. Their average operation counts, compositional-depth proxies, and nearest-training instruction similarities are also similar, indicating that the transfer set does not primarily introduce new task families, rarer operations, or substantially different instruction semantics. Instead, the main shift lies in the structure of the input workbooks.

Compared with the IID set, the structural-transfer set contains more multi-sheet workbooks ($12/40$ versus $5/40$), more workbooks with three or more sheets ($3/40$ versus $0/40$), more workbooks containing existing cross-sheet formulas ($8/40$ versus $1/40$), more tasks requiring cross-sheet reasoning ($13/40$ versus $6/40$), and more merged-cell workbooks ($8/40$ versus $2/40$). The mean number of sheets increases from $1.125$ to $1.475$, and the mean nearest-training workbook-structure distance increases from $0.882$ to $1.676$. We therefore use this set to evaluate transfer across workbook structure rather than transfer to unseen semantic task families.

\begin{table}[H]
\centering
\small
\caption{Characteristics of the two held-out SpreadsheetBench pilot sets. The two sets have closely matched task semantics, while the transfer set contains greater workbook-level structural variation.}
\label{tab:motivating-pilot-split}
\begin{tabular}{lcc}
\toprule
\textbf{Characteristic} & \textbf{IID} & \textbf{Structural transfer} \\
\midrule
Multi-sheet workbooks & 5/40 & 12/40 \\
Three-or-more sheets & 0/40 & 3/40 \\
Existing cross-sheet formulas & 1/40 & 8/40 \\
Cross-sheet requirements & 6/40 & 13/40 \\
Merged-cell workbooks & 2/40 & 8/40 \\
Mean number of sheets & 1.125 & 1.475 \\
Nearest-Train structural distance & 0.882 & 1.676 \\
\bottomrule
\end{tabular}
\end{table}

\paragraph{Protocol.}
The frozen pilot contains 80 training tasks, 40 validation tasks, 40 IID-test tasks, and 40 structural-transfer tasks, with a further 200 SpreadsheetBench tasks excluded from this study. SkillOpt is run for eight epochs with seed 41 and a batch size of 20, yielding 32 skill updates. Intermediate validation-based stopping and other dynamic regularization mechanisms are disabled so that the complete native evolution trajectory can be observed.

We retain the initial skill and the final checkpoint of every epoch. Each checkpoint is evaluated on the same frozen training, validation, IID-test, and structural-transfer sets. The complete checkpoint trajectory is evaluated in three independent evaluation replays, and we report the mean across replays. These replays evaluate the same learned trajectory and therefore should not be interpreted as independent training runs.

\begin{table}[t]
\centering
\small
\caption{Checkpoint performance in the motivating SpreadsheetBench pilot. Values are mean task accuracy across three independent evaluation replays of the same learned trajectory.}
\label{tab:motivating-pilot-performance}
\begin{tabular}{lrrrrr}
\toprule
\textbf{Checkpoint} & \textbf{Train} & \textbf{Val.} & \textbf{IID} & \textbf{Transfer} & \textbf{IID--Transfer} \\
\midrule
Initial & 40.00 & 28.33 & 50.83 & 12.50 & 38.33 \\
E1      & 61.67 & 65.83 & 63.33 & 35.83 & 27.50 \\
E2      & 54.58 & 55.83 & 59.17 & 39.17 & 20.00 \\
E3      & 56.25 & 55.83 & 60.00 & 40.00 & 20.00 \\
E4      & 60.42 & 54.17 & 54.17 & 46.67 & 7.50 \\
E5      & 57.92 & 53.33 & 50.00 & 39.17 & 10.83 \\
E6      & 60.42 & 55.83 & 54.17 & 40.00 & 14.17 \\
E7      & 58.33 & 54.17 & 58.33 & 35.00 & 23.33 \\
E8      & 61.67 & 59.17 & 61.67 & 37.50 & 24.17 \\
\bottomrule
\end{tabular}
\end{table}

\paragraph{Post-peak transfer behavior.}
Structural-transfer performance reaches its maximum of $46.67\%$ at E4. From E4 to E8, train, validation, and IID accuracy change by $+1.25$, $+5.00$, and $+7.50$ percentage points, respectively, whereas transfer accuracy decreases by $9.17$ points. The IID--transfer gap consequently widens from $7.50$ to $24.17$ percentage points. The decline is also distributed across tasks: among the 40 transfer tasks, 13 regress from E4 to E8, 22 remain unchanged, and 5 improve.

\paragraph{Deterministic lexical diagnostics.}
To characterize how the skill changes along the same trajectory without relying on an LLM judge, we apply four deterministic lexical diagnostics to every checkpoint. These measures are descriptive proxies used only in this motivating study and are distinct from the operational complexity measure $C(S)$ used by \method in Appendix~\ref{app:skillopt-measurement}.

\emph{Lexical length} counts word-like sequences, numeric tokens, and non-whitespace punctuation over the full skill document. \emph{Strict-directive lines} count substantive lines containing directive expressions such as \texttt{must}, \texttt{always}, \texttt{never}, or \texttt{do not}. \emph{Conditional-rule lines} count substantive lines containing conditional expressions such as \texttt{if}, \texttt{when}, \texttt{unless}, or \texttt{fallback}. \emph{Concrete-artifact matches} count spreadsheet-specific cell or range references, error identifiers, selected code identifiers, spreadsheet-specific phrases, and selected functions. These diagnostics characterize observable textual structure rather than semantic quality.

\begin{table}[t]
\centering
\small
\caption{Deterministic lexical diagnostics along the motivating pilot trajectory. These are descriptive proxies and are distinct from the complexity measure $C(S)$ used by \method.}
\label{tab:motivating-pilot-specialization}
\begin{tabular}{lrrrr}
\toprule
\textbf{Checkpoint} & \textbf{Lexical length} & \textbf{Strict lines} & \textbf{Conditional lines} & \textbf{Artifact matches} \\
\midrule
Initial & 402   & 4  & 1  & 13  \\
E1      & 1,944 & 12 & 17 & 44  \\
E2      & 3,252 & 22 & 34 & 71  \\
E3      & 4,416 & 31 & 41 & 84  \\
E4      & 5,586 & 40 & 50 & 95  \\
E5      & 6,655 & 50 & 60 & 113 \\
E6      & 7,728 & 60 & 70 & 119 \\
E7      & 8,501 & 67 & 76 & 127 \\
E8      & 9,494 & 78 & 85 & 139 \\
\bottomrule
\end{tabular}
\end{table}

Between the transfer-performance peak at E4 and the final checkpoint E8, lexical length increases by $69.9\%$ and conditional-rule lines increase by $70.0\%$. Over the full trajectory, lexical length, strict-directive lines, conditional-rule lines, and concrete-artifact matches reach $23.6\times$, $19.5\times$, $85\times$, and $10.7\times$ their initial values, respectively.

\paragraph{Interpretation.}
Taken together, the pilot illustrates a post-peak transfer pattern: after structural-transfer performance peaks, continued skill evolution is accompanied by a widening IID--transfer gap and continued accumulation of increasingly specific skill structure. This pattern motivates treating repeated skill updates as a regularized learning process rather than relying on the updater alone.

\subsection{SkillOpt Measurement Conventions and Complexity}
\label{app:skillopt-measurement}

We evaluate SpreadsheetBench ($100/100/200$ train/validation/test examples), SearchQA ($400/200/1400$), and LiveMathBench ($100/80/217$). All SkillOpt runs use eight evolution epochs, eight-example reflection minibatches, and an initial edit budget of four. The native baseline retains SkillOpt's slow-update and meta-skill mechanisms; the \method integration retains the same ranked-edit proposal mechanism and applies regularization before the native acceptance decision.

Final checkpoints are evaluated after the fixed eight-epoch training budget.
Delivery selects among the eight epoch-end snapshots using the incumbent
validation score recorded at the last training update of each epoch, breaking
ties in favor of the earlier epoch. The selected epochs for SkillOpt/\method
are 7/7 on SpreadsheetBench, 4/2 on SearchQA, and 5/1 on LiveMath. Their
selection scores are $59.5/63.5\%$, $80.25/80.25\%$, and
$43.125/38.125\%$, respectively. Test scores average two frozen evaluations of
the exact selected snapshot.

For completeness, SearchQA token-overlap F1 is $87.98/88.68\%$ at delivery
and $88.26/88.41\%$ at the final checkpoint for SkillOpt/\method,
respectively, averaged over the same two frozen evaluations as EM.

Skill length uses the \texttt{o200k\_base} tokenizer on the full document. We
segment skill documents into deterministic exact-span atoms and define
\begin{equation}
C(S)=\sum_{a_i\in\mathcal{A}(S)}
\left(\frac{\tau_i}{100}+0.2+\frac{L_i}{20}
+2\frac{B_i}{5}+2\frac{D_i}{3}\right),
\label{eq:skill-complexity}
\end{equation}
where $\tau_i$, $L_i$, $B_i$, and $D_i$ denote the lexical-unit count,
nonempty code-line count, control-flow count, and external-dependency count of
atom $a_i$, respectively; code terms are zero for non-code atoms. The
exact-span segmenter retains document scaffolding, and each returned atom
receives the base cost. For code, $L_i$ excludes empty lines and Markdown fence
delimiters, $B_i$ counts common control-flow constructs, and $D_i$ counts
recognized imports and selected HTTP dependency calls. The measure is an
operational regularization signal rather than a semantic-quality score.

\begin{table}[t]
\centering
\small
\caption{Final composite complexity $C$ on the SkillOpt benchmarks.}
\label{tab:skillopt-complexity}
\begin{tabular}{clrr}
\toprule
\textbf{Dataset} & \textbf{Method} & \multicolumn{2}{c}{\textbf{Complexity $C$ $\downarrow$}} \\
\cmidrule(lr){3-4}
& & \textbf{Value} & \textbf{Relative to baseline} \\
\midrule
\multirow[c]{2}{*}{SpreadsheetBench}
& SkillOpt & 116.96 & 1.00 \\
& \ourscell{\method} & \ourscell{\textbf{57.21}} & \ourscell{\textbf{0.49}} \\
\midrule
\multirow[c]{2}{*}{SearchQA}
& SkillOpt & 124.42 & 1.00 \\
& \ourscell{\method} & \ourscell{\textbf{39.51}} & \ourscell{\textbf{0.32}} \\
\midrule
\multirow[c]{2}{*}{LiveMath}
& SkillOpt & 110.52 & 1.00 \\
& \ourscell{\method} & \ourscell{\textbf{36.52}} & \ourscell{\textbf{0.33}} \\
\bottomrule
\end{tabular}
\end{table}

\subsection{Evaluation Protocol Conventions}
\label{app:evaluation-conventions}

All reported performance values use the benchmark-native verifier and the same task order for compared methods. Frozen evaluations do not update the skill state. When a benchmark provides a validation stream, checkpoint selection uses only the recorded validation scores and the selected checkpoint is then evaluated on the frozen test stream. Main comparisons use one evolved state per condition; repeated frozen evaluations measure execution-time variation of that state rather than variation across independently trained trajectories.

\section{Framework-Specific Instantiations of \method}
\label{app:framework-instantiations}

\method is a general regularization framework rather than a single shared software core or a fixed pipeline implementation. The three benchmark integrations retain their native skill evolvers and task evaluators but differ in skill representation, complexity semantics, CCV case construction, repair scope, and final candidate handling.

\begin{table}[H]
\centering
\footnotesize
\caption{Framework-specific integration of \method.}
\label{tab:framework-instantiations}
\setlength{\tabcolsep}{3pt}
\begin{tabular}{lp{0.23\linewidth}p{0.30\linewidth}p{0.29\linewidth}}
\toprule
\textbf{Framework} & \textbf{Native proposal} & \textbf{Complexity handling} & \textbf{Final candidate handling} \\
\midrule
SkillOpt &
Ranked edits to one instruction document &
Complexity contributes to the regularized validation gate &
Regularized acceptance gate; current/best checkpoint bookkeeping retained \\
\midrule
SkillEvol &
\textsc{SkillAuthor} patch over a multi-skill family &
Growth budget can trigger bounded local regularization; budget alone does not reject &
Native library commit API reused with staging, optional repair, and finalization \\
\midrule
Continual &
Native \textsc{Create-Skill}/\textsc{Modify-Skill} over a skill directory &
CREATE uses an absolute budget and library-level comparison; MODIFY uses relative growth &
No validation reject of the final regularized candidate; final snapshot is written back \\
\bottomrule
\end{tabular}
\end{table}

\begin{table}[H]
\centering
\footnotesize
\caption{Framework-specific CCV instantiations used in our experiments.}
\label{tab:ccv-instantiations}
\setlength{\tabcolsep}{3pt}
\begin{tabular}{@{}lp{0.27\linewidth}p{0.34\linewidth}p{0.19\linewidth}@{}}
\toprule
\textbf{Framework} & \textbf{CCV case construction} & \textbf{Regression signal} & \textbf{Repair scope} \\
\midrule
SkillOpt &
Dataset-specific candidate-conditioned attacks over workbook, context, or reasoning inputs &
Hard incumbent pass and candidate failure, with clean-candidate and dataset-specific attribution controls &
Candidate delta and authorized related atoms \\
\midrule
SkillEvol &
Original instruction plus a bounded robustness appendix &
Qualified clean pair, old attacked score $\geq 0.8$, and attacked score drop $\geq 0.05$ &
Changed skill and implicated local regions \\
\midrule
Continual &
Answer-preserving transformation or validated boundary case &
Incumbent passes and candidate either fails or drops by more than $0.1$ &
Responsible skill paths \\
\bottomrule
\end{tabular}
\end{table}

Across all three integrations, the incumbent and candidate must first satisfy the framework-specific clean qualification, after which both are evaluated on the same attacked case under the same model configuration and benchmark-native evaluator. SkillOpt uses a strict hard-pass/hard-fail attacked-case criterion with additional attribution controls, whereas SkillEvolBench and ContinualSkillBench also allow a framework-defined continuous-score drop to trigger a regression signal.

\section{Additional Benchmark Results}
\label{app:additional-results}

\subsection{LiveMath: A Distribution-Sensitive Answer Heuristic}
\label{app:livemath-shortcut}

LiveMath provides a case in which suppressing repetitive specialization can trade off against accuracy on the current test distribution. SkillOpt's final skill repeatedly emphasizes the meta-option stating that a stronger result can be proved. The regularized final skill retains one explicit meta-option rule but consolidates overlapping strongest-statement and strongest-existence guidance.

Of the 217 test questions, 92 have the correct answer text ``One of the remaining options is correct, but a stronger result can be proven.'' Across the two frozen evaluation repeats, \method's delivery checkpoint answers 33 of 184 corresponding question--repeat instances correctly, whereas the final checkpoint answers 15. The remaining 125 questions change only from 119/250 to 116/250 correct. Thus, meta-option questions account for 18 of the 21 net lost correct answers: $4.15$ of the overall $4.84$ percentage-point decline, or $85.7\%$. Pairing delivery and final outcomes within each repeat yields 23 correct-to-incorrect transitions and five reverse transitions on this subset.

The concentration is consistent with reduced reinforcement of the meta-option heuristic, but the comparison does not isolate any individual rule edit as causal. It illustrates that regularization can suppress repetitive specialization even when the specialized pattern remains useful under the current evaluation distribution. The case therefore illustrates a genuine regularization trade-off: suppressing repeated specialization can reduce benchmark accuracy when the evaluation distribution itself rewards the specialized answer prior.

\subsection{SkillEvolBench Measurements and Diagnostics}
\label{app:skillevol-measurement}

SkillEvolBench contains six environments with five task families each; T1--T3 are acquisition tasks and T4--T6 are deployment tasks. In the original benchmark protocol, the acquisition sequence is traversed once before the resulting skill library is evaluated on unseen deployment tasks. Our 1-pass condition corresponds to this native acquisition endpoint.

To study later-stage skill evolution after a library has already formed, we introduce a controlled second acquisition pass; this 2-pass setting is not part of the original SkillEvolBench protocol. Both the unregularized 2-pass condition and \method start from the exact same one-pass library and process the same T1--T3 acquisition sequence again in the same order. Thus, they receive identical additional experience, while T4--T6 remain unseen throughout acquisition; the comparison isolates how that repeated experience is incorporated into the existing skill library. \method retains the native execution environment, verifier, SkillAuthor proposal, and library commit interface, inserting regularization only around the candidate update before write-back. Deployment and replay evaluations use frozen libraries and two independent sessions.

Library tokens use \texttt{o200k\_base}, summed over the 32 final
\texttt{SKILL.md} files across six isolated environments. Structural atoms use
the same deterministic exact-span segmentation as
Equation~\ref{eq:skill-complexity}. We sum $C$ over individual files and report
growth relative to the one-pass library. The original benchmark-specific atom counts
are 991, 1,113, and 1,070 for the one-pass, two-pass, and regularized libraries;
the common structural counts are 1,422, 1,551, and 1,515, respectively.
Relative complexity growth is $14.32\%$ for two passes and $9.65\%$ with
\method.

\subsubsection{Continuous verifier scores}
\label{app:skillevol-continuous}

In addition to strict success, we report the benchmark's normalized overall
verifier score, which gives partial credit according to each task's official
rubric. We average the two frozen evaluations of each task and then average
across tasks in each block.

\begin{table}[t]
\centering
\small
\caption{Mean normalized SkillEvolBench verifier scores, averaged over two
frozen evaluations. Scores range from 0 to 1.}
\label{tab:skillevol-continuous}
\begin{tabular}{lrr>{\columncolor{methodshade}}r}
\toprule
\textbf{Task block} & \textbf{1-pass} & \textbf{2-pass} & \textbf{\method} \\
\midrule
T1--T3, Replay       & 0.8557 & \textbf{0.8588} & 0.8558 \\
\quad T1, Canonical replay & 0.8604 & 0.8625 & \textbf{0.8807} \\
\quad T2, Enriched replay  & \textbf{0.8484} & 0.8439 & 0.8381 \\
\quad T3, Variant replay   & 0.8584 & \textbf{0.8699} & 0.8487 \\
T4--T6, Deployment   & \textbf{0.8299} & 0.8263 & 0.8269 \\
\quad T4, Context shift    & 0.8334 & \textbf{0.8447} & 0.8227 \\
\quad T5, Adversarial      & \textbf{0.8317} & 0.8226 & 0.8297 \\
\quad T6, Composition      & 0.8247 & 0.8115 & \textbf{0.8283} \\
Complete, T1--T6     & \textbf{0.8428} & 0.8425 & 0.8414 \\
\bottomrule
\end{tabular}
\end{table}

The deployment means differ by less than 0.004 despite larger differences in
strict success. The two composition evaluations yield 6/30 and 5/30 successes
for one pass, 4/30 and 4/30 for two passes, and 8/30 and 8/30 with \method.
All conditions retrieve the required skills on all original composition tasks,
making missing-skill retrieval an unlikely explanation for the composition
difference.

\subsection{ContinualSkillBench Measurements and Complete Results}
\label{app:continual-measurement}

ContinualSkillBench differs from the other settings in that its native updater can expand the skill library through explicit \textsc{Create-Skill} actions in addition to revising existing skills through \textsc{Modify-Skill}. The evolving state therefore includes both the contents of individual skills and the library structure itself. We consequently treat library proliferation as part of the regularization problem: CREATE introduces a new persistent skill, whereas MODIFY changes an existing one.

Training processes tasks 1--50 sequentially while retaining the evolving library. Frozen evaluation then uses the step-50 library on tasks 51--100, with a fresh session for each task and no further skill updates. Baseline and \method share the same task order, model, and evaluation procedure. The two evaluation repeats use the same frozen library and therefore measure evaluation variation rather than variation across training runs.

\paragraph{Online learning performance.}
Table~\ref{tab:continual-learning50} follows the source benchmark's reporting
format and evaluates tasks 1--50 as they are encountered during skill
evolution. Raw is the mean over all 50 tasks; Norm. is the mean over tasks with
scoreable outputs in both conditions. All 50 tasks are common-valid in every
domain here, so Raw and Norm. coincide.

\begin{table}[t]
\centering
\small
\caption{ContinualSkillBench online learning performance on tasks 1--50,
reported in the source benchmark's format. Entries are rewards in $[0,1]$ from
a single sequential learning trajectory.}
\label{tab:continual-learning50}
\setlength{\tabcolsep}{3.5pt}
\resizebox{\linewidth}{!}{%
\begin{tabular}{clrrrrrrr}
\toprule
\textbf{Domain} & \textbf{Method} & \textbf{EM} & \textbf{F1} & \textbf{Num.} & \textbf{Prog.} & \textbf{Rubric} & \textbf{Raw} & \textbf{Norm.} \\
\midrule
\multirow[c]{2}{*}{Mathematics}
& Baseline & \textbf{0.641} & -- & \textbf{0.909} & -- & -- & \textbf{0.700} & \textbf{0.700} \\
& \ourscell{\method} & \ourscell{0.615} & \ourscell{--} & \ourscell{0.818} & \ourscell{--} & \ourscell{--} & \ourscell{0.660} & \ourscell{0.660} \\
\midrule
\multirow[c]{2}{*}{Law}
& Baseline & \textbf{0.853} & -- & -- & \textbf{0.800} & 0.615 & 0.795 & 0.795 \\
& \ourscell{\method} & \ourscell{0.824} & \ourscell{--} & \ourscell{--} & \ourscell{\textbf{0.800}} & \ourscell{\textbf{0.774}} & \ourscell{\textbf{0.810}} & \ourscell{\textbf{0.810}} \\
\midrule
\multirow[c]{2}{*}{Finance}
& Baseline & 0.364 & \textbf{0.488} & \textbf{0.583} & \textbf{1.000} & \textbf{0.513} & \textbf{0.541} & \textbf{0.541} \\
& \ourscell{\method} & \ourscell{\textbf{0.545}} & \ourscell{0.392} & \ourscell{0.417} & \ourscell{0.600} & \ourscell{0.492} & \ourscell{0.478} & \ourscell{0.478} \\
\midrule
\multirow[c]{2}{*}{Office}
& Baseline & -- & -- & 0.862 & \textbf{0.933} & \textbf{0.913} & 0.890 & 0.890 \\
& \ourscell{\method} & \ourscell{--} & \ourscell{--} & \ourscell{\textbf{0.897}} & \ourscell{\textbf{0.933}} & \ourscell{0.908} & \ourscell{\textbf{0.909}} & \ourscell{\textbf{0.909}} \\
\midrule
\multirow[c]{2}{*}{Healthcare}
& Baseline & 0.750 & -- & -- & \textbf{0.500} & 0.759 & 0.736 & 0.736 \\
& \ourscell{\method} & \ourscell{\textbf{0.813}} & \ourscell{--} & \ourscell{--} & \ourscell{\textbf{0.500}} & \ourscell{\textbf{0.769}} & \ourscell{\textbf{0.761}} & \ourscell{\textbf{0.761}} \\
\bottomrule
\end{tabular}}
\end{table}

\paragraph{Held-out evaluator breakdown.}
Structured in Table~\ref{tab:exp-continual} averages individual non-rubric task
scores across both repeats rather than averaging evaluator-category means
equally. The source-category breakdown is shown in
Table~\ref{tab:continual-types}.

\begin{table}[t]
\centering
\small
\caption{Mean frozen reward by the source benchmark's evaluator categories.
Every score averages two repeats.}
\label{tab:continual-types}
\setlength{\tabcolsep}{4pt}
\begin{tabular}{clrrrrr}
\toprule
\textbf{Domain} & \textbf{Method} & \textbf{EM} & \textbf{F1} & \textbf{Num.} & \textbf{Prog.} & \textbf{Rubric} \\
\midrule
\multirow[c]{2}{*}{Mathematics}
& Baseline & -- & -- & 73.08 & 75.00 & 61.80 \\
& \ourscell{\method} & \ourscell{--} & \ourscell{--} & \ourscell{73.08} & \ourscell{75.00} & \ourscell{\textbf{65.00}} \\
\midrule
\multirow[c]{2}{*}{Law}
& Baseline & 100.00 & -- & -- & 60.00 & 60.79 \\
& \ourscell{\method} & \ourscell{100.00} & \ourscell{--} & \ourscell{--} & \ourscell{\textbf{70.00}} & \ourscell{\textbf{66.27}} \\
\midrule
\multirow[c]{2}{*}{Finance}
& Baseline & -- & 21.73 & -- & 87.50 & 69.59 \\
& \ourscell{\method} & \ourscell{--} & \ourscell{\textbf{22.86}} & \ourscell{--} & \ourscell{87.50} & \ourscell{\textbf{70.09}} \\
\midrule
\multirow[c]{2}{*}{Office}
& Baseline & 100.00 & -- & \textbf{75.00} & 80.00 & \textbf{85.33} \\
& \ourscell{\method} & \ourscell{100.00} & \ourscell{--} & \ourscell{71.43} & \ourscell{80.00} & \ourscell{84.57} \\
\midrule
\multirow[c]{2}{*}{Healthcare}
& Baseline & \textbf{66.67} & -- & -- & -- & 72.27 \\
& \ourscell{\method} & \ourscell{62.50} & \ourscell{--} & \ourscell{--} & \ourscell{--} & \ourscell{\textbf{76.47}} \\
\bottomrule
\end{tabular}
\end{table}

\paragraph{CREATE/MODIFY regularization.}
The two native update modes induce different growth semantics. For \textsc{Create-Skill}, \method controls the absolute complexity of the newly introduced skill and considers overlap with the existing library; for \textsc{Modify-Skill}, it regularizes the candidate relative to the incumbent skill and its update-induced growth. In both cases, the final regularized state is written back through the benchmark's native continual-learning workflow.

\paragraph{Library measurements.}
Skill count is the number of skill directories containing a
\texttt{SKILL.md} in the delivered domain library, including retained initial
domain skills. Tokens use \texttt{o200k\_base} and are summed over skill
documents and bundled auxiliary text/code files. Complexity uses
Equation~\ref{eq:skill-complexity}; auxiliary files contribute lexical length
and code cost. All measurements use the step-50 frozen snapshot.

\paragraph{Office sensitivity analysis.}
Office's $1.44$-point Overall difference is concentrated rather than broad.
All but three of the 21 structured tasks are identical between conditions. One
rubric task additionally generates a complete briefing but writes it to a path
different from the evaluated path. We retain the strict end-to-end score in
the main table. Excluding this single task reduces the Overall gap to $0.63$
points; excluding it together with the three table-selection outliers leaves
\method $0.42$ points higher on the other 46 tasks. We use this analysis only
to localize the observed difference, not to replace the official score.

\paragraph{Healthcare evaluator analysis.}
Healthcare gains $2.20$ Overall points while reducing the library from 42 to
34 skills. Its Rubric score improves by $4.21$ points, whereas the Structured
difference corresponds to one net exact-match success across the 24
question--repeat instances. The aggregate gain therefore comes primarily from
open-ended transfer rather than from easier exact-match behavior.

\subsection{Deployment Inference Usage and Skill Compactness}
\label{app:inference-cost}

Inference usage is not an optimization target of \method; we examine it as a deployment-side by-product of compacting the persistent skill state. A shorter persistent skill can reduce the context consumed during downstream execution, but end-to-end model usage also depends on the execution trajectory, particularly when external search, document retrieval, or other tool interactions introduce substantial and variable context. For each evaluation run, we count input plus output tokens from the completed model session, including cached input. Judge calls, training, regularizer calls, and superseded failed attempts are excluded. Every row uses the frozen evaluation segment corresponding to the reported checkpoint and pools the two independent repeats.

\begin{table}[H]
\centering
\small
\caption{Frozen-evaluation model-token usage. Values are total input-plus-output tokens per task, pooled over two repeats; trimmed means remove the top and bottom $10\%$ of tasks.}
\label{tab:continual-inference-tokens}
\setlength{\tabcolsep}{4pt}
\begin{tabular}{clrrr}
\toprule
\textbf{Benchmark} & \textbf{Method} & \textbf{Mean} & \textbf{Median} & \textbf{Trimmed mean} \\
\midrule
\multirow[c]{2}{*}{SpreadsheetBench}
& SkillOpt & 12,572 & 10,099 & 11,266 \\
& \ourscell{\method} & \ourscell{\textbf{8,865}} & \ourscell{\textbf{7,328}} & \ourscell{\textbf{7,880}} \\
\midrule
\multirow[c]{2}{*}{SearchQA}
& SkillOpt & 6,280 & 6,245 & 6,257 \\
& \ourscell{\method} & \ourscell{\textbf{2,759}} & \ourscell{\textbf{2,713}} & \ourscell{\textbf{2,726}} \\
\midrule
\multirow[c]{2}{*}{LiveMath}
& SkillOpt & 8,186 & 7,426 & 7,789 \\
& \ourscell{\method} & \ourscell{\textbf{5,028}} & \ourscell{\textbf{4,437}} & \ourscell{\textbf{4,637}} \\
\midrule
\multirow[c]{2}{*}{SkillEvolBench}
& 2-pass & 396,513 & 334,137 & 353,599 \\
& \ourscell{\method} & \ourscell{\textbf{359,340}} & \ourscell{\textbf{281,948}} & \ourscell{\textbf{323,701}} \\
\midrule
\multirow[c]{2}{*}{Continual--Math}
& Baseline & \textbf{153,611} & 74,418 & \textbf{113,380} \\
& \ourscell{\method} & \ourscell{168,458} & \ourscell{\textbf{73,441}} & \ourscell{105,502} \\
\midrule
\multirow[c]{2}{*}{Continual--Law}
& Baseline & \textbf{401,321} & 277,371 & \textbf{336,404} \\
& \ourscell{\method} & \ourscell{412,642} & \ourscell{\textbf{260,898}} & \ourscell{347,242} \\
\midrule
\multirow[c]{2}{*}{Continual--Finance}
& Baseline & 1,236,906 & 418,529 & 694,990 \\
& \ourscell{\method} & \ourscell{\textbf{863,415}} & \ourscell{\textbf{256,081}} & \ourscell{\textbf{558,126}} \\
\midrule
\multirow[c]{2}{*}{Continual--Office}
& Baseline & 238,229 & 129,351 & 155,029 \\
& \ourscell{\method} & \ourscell{\textbf{220,137}} & \ourscell{\textbf{114,689}} & \ourscell{\textbf{130,362}} \\
\midrule
\multirow[c]{2}{*}{Continual--Healthcare}
& Baseline & \textbf{156,572} & \textbf{121,384} & \textbf{140,291} \\
& \ourscell{\method} & \ourscell{194,875} & \ourscell{129,183} & \ourscell{150,549} \\
\bottomrule
\end{tabular}
\end{table}

The main pattern is that skill compactness reduces a stable component of inference context, but whether this translates into lower end-to-end token usage depends on how much execution is dominated by trajectory-dependent external context. The effect is clearest on tasks with relatively controlled execution paths. On SpreadsheetBench and LiveMath, where task context is fixed and execution does not depend on open-ended external information gathering, \method reduces mean frozen-evaluation token usage by $29.5\%$ and $38.6\%$, respectively, with corresponding reductions in both the median and trimmed mean. SearchQA also shows a substantial $56.1\%$ reduction despite using web search, indicating that external search does not by itself eliminate the benefit of a compact skill state. SkillEvolBench likewise reduces mean usage by $9.4\%$ relative to the matched two-pass condition. Together, these results show that compacting the persistent skill can produce substantial deployment-side savings when the skill state remains a meaningful and comparatively stable component of the inference context.

ContinualSkillBench exposes a different regime, in which high-variance information-acquisition and tool-use trajectories can dominate total inference usage. Finance and Office decrease across the principal summaries, whereas Healthcare increases; Mathematics and Law have higher arithmetic means under \method despite lower medians, indicating that a small number of long trajectories can dominate average usage. To examine this pattern more closely, we pair the same frozen-evaluation task and repeat under the baseline and \method for Mathematics, Law, and Healthcare. As one measurable indicator of externally mediated trajectories, we separate pairs according to whether either execution invokes \texttt{web\_search}. Across the resulting 300 pairs, the three domains increase by 6.45 million tokens in aggregate; pairs involving \texttt{web\_search} account for a 6.66 million-token increase, whereas pairs with no \texttt{web\_search} decrease by 0.21 million tokens overall. The contrast is particularly pronounced in Mathematics and Healthcare, where no-search pairs decrease by 6,754 and 1,005 tokens per task on average, respectively, while search-associated pairs increase by 53,247 and 73,161 tokens. Law is more mixed: search-associated pairs contribute most of the net increase, but several high-cost trajectories involve extensive local-document processing without web search.

The \texttt{web\_search} split should therefore be interpreted as an observable indicator of externally mediated trajectories rather than as a causal attribution to search itself. High-usage trajectories can also involve substantially different amounts of PDF or document extraction, shell-returned content, and prolonged tool interaction, while the available logs do not expose the exact token contribution of each source after context injection and truncation. The resulting picture is therefore not that compact skills universally reduce total inference tokens, but that they reduce one persistent and controllable source of context consumption. When execution context is relatively stable, this reduction can translate into substantial token savings; when external context dominates the trajectory, the same benefit can be obscured by much larger variation in tool-mediated context.

\paragraph{Takeaway.} Reduced inference usage is a deployment-side by-product rather than an optimization objective of \method. Compact persistent skills reduce the controllable skill-context component of inference usage, while end-to-end savings depend on how strongly the downstream trajectory is dominated by external information and tool interactions.

\section{Algorithmic Details and Updater Integration}
\label{app:regularization-algorithms}

This appendix formalizes the three regularizers introduced in Section~\ref{sec:method} and describes how they connect to different native skill updaters. The regularization roles are shared across SkillOpt, SkillEvolBench, and ContinualSkillBench, while the update boundary is framework-dependent: candidate changes may be represented as document edits, localized library patches, or changes between skill-library snapshots.

\subsection{Native Update Interface}
\label{app:native-update-interface}

We represent one native evolution step abstractly as
\begin{equation}
    (\tau_t,\widetilde{S}_{t+1})
    =
    \operatorname{NativeUpdate}(S_t,x_t),
    \label{eq:native-update-interface}
\end{equation}
where $S_t$ is the incumbent skill state, $x_t$ is the current training task or batch, $\tau_t$ is the resulting execution trajectory, and $\widetilde{S}_{t+1}$ is the native candidate. When dropout is active, the native update is first generated from a temporary masked view and then reconciled with the complete incumbent, as described below.

Throughout the appendix, $\operatorname{Diff}(S,S')$ denotes the change from a source state $S$ to a candidate state $S'$. We use stage-specific superscripts for the resulting deltas: $\Delta_t^{\mathrm{drop}}$ for the change learned under the dropout view, $\Delta_t^{\mathrm{comp}}$ for the reconciled candidate change examined by complexity regularization, and $\Delta_t^{\mathrm{ccv}}$ for the post-complexity candidate change examined by CCV. This keeps the underlying notion of a candidate delta shared while distinguishing the state boundary at which it is computed.

The main framework-dependent integration operations are \texttt{Reconcile}, which transfers an update learned from a temporarily perturbed skill view onto the complete incumbent, and \texttt{WriteBack}, which returns the final regularized candidate through the native framework's acceptance or commit interface.

\subsection{Training-Time Skill Dropout}
\label{app:skill-dropout-algorithm}

A \emph{skill atom} is a locally meaningful unit of reusable skill content, such as an individual directive, list item, compact procedural block, or another semantically coherent segment. Atom segmentation provides stable units for temporary masking and later local structural comparison.

Algorithm~\ref{alg:skill-dropout} formalizes training-time skill dropout. Its central invariant is that masking affects update generation only. Content removed from the temporary view is restored before the dropout-stage candidate proceeds to subsequent regularization.

\begin{algorithm}[t]
\caption{Training-Time Skill Dropout}
\label{alg:skill-dropout}
\begin{algorithmic}[1]
\Require Incumbent skill state $S_t$, training task $x_t$
\Ensure Dropout-reconciled candidate $S_{t+1}^{\mathrm{drop}}$
\State $\mathcal{A}_t \gets \operatorname{SegmentAtoms}(S_t)$
\State $M_t \gets \operatorname{SampleMask}(\mathcal{A}_t)$
\State $S_t^{\mathrm{drop}} \gets \operatorname{Mask}(S_t,M_t)$
\State $(\tau_t,\widetilde{S}_{t+1}^{\mathrm{drop}}) \gets \operatorname{NativeUpdate}(S_t^{\mathrm{drop}},x_t)$
\State $\Delta_t^{\mathrm{drop}} \gets \operatorname{Diff}(S_t^{\mathrm{drop}},\widetilde{S}_{t+1}^{\mathrm{drop}})$
\State $S_{t+1}^{\mathrm{drop}} \gets \operatorname{Reconcile}(S_t,\Delta_t^{\mathrm{drop}})$
\State \Return $S_{t+1}^{\mathrm{drop}}$
\end{algorithmic}
\end{algorithm}

\paragraph{Atom segmentation.} For textual skill states, atoms are obtained from deterministic structural segmentation of the skill document, so that repeated references to the same instruction or procedural block remain stable across an update. Multi-skill libraries additionally retain skill-file or skill-path boundaries. These higher-level boundaries determine the region within which atom-level operations are permitted.

\paragraph{Restoration.} Because the native updater observes $S_t^{\mathrm{drop}}$ rather than the complete incumbent $S_t$, $\Delta_t^{\mathrm{drop}}$ captures only the change proposed under the perturbed context. \texttt{Reconcile} restores the masked atoms from $S_t$ and transfers that change onto the complete incumbent, producing $S_{t+1}^{\mathrm{drop}}$. An atom is therefore never removed merely because it was absent from the temporary training-time view.

\subsection{Complexity-Aware Local Regularization}
\label{app:complexity-algorithm}

Complexity-aware local regularization operates on the dropout-reconciled candidate $S_{t+1}^{\mathrm{drop}}$. The stage-local change is
\begin{equation}
    \Delta_t^{\mathrm{comp}}
    =
    \operatorname{Diff}
    \left(
        S_t,
        S_{t+1}^{\mathrm{drop}}
    \right),
\end{equation}
which is measured relative to the complete incumbent, unlike $\Delta_t^{\mathrm{drop}}$ in the preceding stage.

Algorithm~\ref{alg:complexity-regularization} separates complexity measurement, local-scope construction, and discrete edit selection.

\begin{algorithm}[t]
\caption{Complexity-Aware Local Regularization}
\label{alg:complexity-regularization}
\begin{algorithmic}[1]
\Require Incumbent skill state $S_t$, dropout-reconciled candidate $S_{t+1}^{\mathrm{drop}}$
\Ensure Complexity-regularized candidate $S_{t+1}^{\mathrm{comp}}$
\State $\Delta_t^{\mathrm{comp}} \gets \operatorname{Diff}(S_t,S_{t+1}^{\mathrm{drop}})$
\State $c_t \gets f_{\mathrm{comp}}\!\left(\phi(S_t),\phi(S_{t+1}^{\mathrm{drop}}),\phi(\Delta_t^{\mathrm{comp}})\right)$
\State $S_{t+1}^{\mathrm{comp}} \gets S_{t+1}^{\mathrm{drop}}$
\If{$\operatorname{NeedsRegularization}(c_t,\Delta_t^{\mathrm{comp}})$}
    \State $\mathcal{D}_t \gets \operatorname{SegmentAtoms}(\Delta_t^{\mathrm{comp}})$
    \State $\mathcal{R}_t \gets \operatorname{RetrieveRelated}(S_t,\mathcal{D}_t)$
    \State $\Omega_t \gets \operatorname{BoundScope}(\mathcal{D}_t \cup \mathcal{R}_t)$
    \ForAll{$a \in \mathcal{D}_t$}
        \State $o_a \gets \operatorname{SelectOperation}(a,\Omega_t,c_t)$
        \State $S_{t+1}^{\mathrm{comp}} \gets \operatorname{ApplyLocalOperation}(S_{t+1}^{\mathrm{comp}},a,o_a,\Omega_t)$
    \EndFor
\EndIf
\State \Return $S_{t+1}^{\mathrm{comp}}$
\end{algorithmic}
\end{algorithm}

\subsubsection{Complexity Signal Definitions}
\label{app:complexity-definitions}

Equation~\ref{eq:complexity-signal} leaves the structural feature extractor abstract in the main text. We instantiate it as
\begin{equation}
    \phi(Z)
    =
    \big[
        \phi_{\mathrm{size}}(Z),
        \phi_{\mathrm{atom}}(Z),
        \phi_{\mathrm{structure}}(Z),
        \phi_{\mathrm{code}}(Z),
        \ldots
    \big]
\end{equation}
for either a skill state or an update delta. For a skill state, the components summarize lexical, atom-level, structural, and executable-code properties; for an update delta, they summarize the corresponding structural change introduced or removed by the candidate.

The regularization signal is
\begin{equation}
c_t
=
f_{\mathrm{comp}}
\left(
    \phi(S_t),
    \phi(S_{t+1}^{\mathrm{drop}}),
    \phi(\Delta_t^{\mathrm{comp}})
\right).
\end{equation}
The function $f_{\mathrm{comp}}$ need not be identical across native update protocols; it determines how shared structural measurements are converted into the signal used to activate or score regularization.

\paragraph{SkillOpt.}
For SkillOpt, the structural features are summarized by the composite complexity measure $C(S)$ defined in Equation~\ref{eq:skill-complexity}. The corresponding $f_{\mathrm{comp}}$ compares incumbent and candidate complexity and supplies the complexity term used by regularized candidate evaluation. Complexity can therefore influence whether a candidate transition is retained.

\paragraph{SkillEvolBench.}
For SkillEvolBench, $f_{\mathrm{comp}}$ is used primarily as a growth trigger. When the candidate exceeds the configured structural-growth budget, it enters bounded local regularization. The signal determines whether additional consolidation is invoked rather than acting as an independent hard rejection criterion.

\paragraph{ContinualSkillBench.}
ContinualSkillBench conditions $f_{\mathrm{comp}}$ on the native update type. For \textsc{Create}, the signal uses an absolute complexity budget for the newly created skill together with its relation to existing library content. For \textsc{Modify}, it emphasizes relative growth of the modified skill with respect to its pre-update state. These are soft regularization triggers: exceeding them can invoke structural reconsideration without automatically vetoing the final candidate.

\paragraph{Discrete local operations.}
The operation-based formulation is inspired by Mem0's memory reconciliation mechanism~\citep{chhikara2025mem0}, in which newly extracted facts are compared with related memories and assigned one of four actions: \textsc{Add}, \textsc{Update}, \textsc{Delete}, or \textsc{NoOp}. Our problem begins after the native skill updater has already proposed new persistent content, so we adapt the discrete-action principle to post-update skill reconciliation:
\begin{equation}
    o_a
    \in
    \{
    \textsc{NoOp},
    \textsc{Merge},
    \textsc{Rewrite},
    \textsc{Delete}
    \}.
\end{equation}
\textsc{NoOp} retains the candidate atom unchanged. \textsc{Merge} consolidates overlapping candidate and incumbent atoms. \textsc{Rewrite} modifies useful but unnecessarily narrow, redundant, or overspecified content. \textsc{Delete} removes candidate content that contributes no distinct capability.

\paragraph{Related-content retrieval.}
For each candidate atom, the regularizer retrieves a bounded number of semantically or structurally related incumbent atoms. Retrieval defines the comparison neighborhood rather than the edit decision itself. High similarity does not imply that two atoms are interchangeable, since apparently redundant instructions can play distinct operational roles in language-model execution.

\paragraph{Granularity across updaters.}
SkillOpt and SkillEvolBench expose the four operations directly over candidate-relative local content. ContinualSkillBench enforces the same locality principle through affected skill paths and can rewrite the full text of an implicated skill. This changes the editing granularity but not the role of the regularizer.

\paragraph{Complexity semantics.}
The complexity signal can combine textual size, structural fragmentation, executable-code structure, and relative growth. SkillOpt can consume it as part of candidate acceptance; SkillEvolBench uses excessive growth to trigger bounded local regularization; and ContinualSkillBench uses soft absolute and relative budgets for \textsc{Create} and \textsc{Modify}, respectively. These differences determine when local reconsideration is activated rather than defining different regularization mechanisms.

\subsection{CCV Implementation Details}
\label{app:ccv-algorithm}

Section~\ref{subsec:skillevoreg-ccv} presents the four-way behavioral comparison. Here we make the underlying checks explicit. CCV validates the post-complexity transition from $S_t$ to $S_{t+1}^{\mathrm{comp}}$, so its stage-local change is $\Delta_t^{\mathrm{ccv}}=\operatorname{Diff}(S_t,S_{t+1}^{\mathrm{comp}})$. This is distinct from $\Delta_t^{\mathrm{comp}}$, which is computed before structural regularization.

\begin{algorithm}[t]
\caption{Causal Counterexample Validation}
\label{alg:ccv}
\begin{algorithmic}[1]
\Require Incumbent skill state $S_t$, complexity-regularized candidate $S_{t+1}^{\mathrm{comp}}$, training history $\mathcal{H}_t$, CCV configuration $\theta_{\mathrm{ccv}}$
\Ensure CCV-processed candidate $S_{t+1}^{\mathrm{ccv}}$
\State $S_{t+1}^{\mathrm{ccv}} \gets S_{t+1}^{\mathrm{comp}}$
\State $\Delta_t^{\mathrm{ccv}} \gets \operatorname{Diff}(S_t,S_{t+1}^{\mathrm{comp}})$
\State $e_t \gets \operatorname{SelectSource}(\mathcal{H}_t,S_t,S_{t+1}^{\mathrm{comp}})$
\If{$e_t=\varnothing$}
    \State \Return $S_{t+1}^{\mathrm{ccv}}$
\EndIf
\State $r_{\mathrm{old}}^{\mathrm{clean}} \gets \operatorname{Evaluate}(S_t,e_t)$
\State $r_{\mathrm{new}}^{\mathrm{clean}} \gets \operatorname{Evaluate}(S_{t+1}^{\mathrm{comp}},e_t)$
\State $q_t \gets \operatorname{CleanQualify}(r_{\mathrm{old}}^{\mathrm{clean}},r_{\mathrm{new}}^{\mathrm{clean}})$
\If{$q_t=0$}
    \State \Return $S_{t+1}^{\mathrm{ccv}}$
\EndIf
\State $a_t \gets \operatorname{GenerateAttack}(e_t,\Delta_t^{\mathrm{ccv}})$
\If{$a_t=\varnothing$}
    \State \Return $S_{t+1}^{\mathrm{ccv}}$
\EndIf
\State $x_t^{\mathrm{ccv}} \gets \operatorname{ApplyAttack}(e_t,a_t)$
\If{$x_t^{\mathrm{ccv}}=\varnothing$}
    \State \Return $S_{t+1}^{\mathrm{ccv}}$
\EndIf
\State $r_{\mathrm{old}}^{\mathrm{attack}} \gets \operatorname{Evaluate}(S_t,x_t^{\mathrm{ccv}})$
\State $r_{\mathrm{new}}^{\mathrm{attack}} \gets \operatorname{Evaluate}(S_{t+1}^{\mathrm{comp}},x_t^{\mathrm{ccv}})$
\State $g_t \gets \operatorname{RegressionCriterion}(r_{\mathrm{old}}^{\mathrm{attack}},r_{\mathrm{new}}^{\mathrm{attack}};\theta_{\mathrm{ccv}})$
\If{$g_t=1$}
    \State $\Omega_t^{\mathrm{ccv}} \gets \operatorname{ImplicatedScope}(\Delta_t^{\mathrm{ccv}},a_t,r_{\mathrm{old}}^{\mathrm{attack}},r_{\mathrm{new}}^{\mathrm{attack}})$
    \State $S_{t+1}^{\mathrm{ccv}} \gets \operatorname{Repair}(S_{t+1}^{\mathrm{comp}},\Omega_t^{\mathrm{ccv}})$
\EndIf
\State \Return $S_{t+1}^{\mathrm{ccv}}$
\end{algorithmic}
\end{algorithm}

\paragraph{Source selection and clean qualification.}
CCV draws $e_t$ from previously observed training trajectories rather than held-out evaluation data. The first two checks correspond to the first column of Equation~\ref{eq:ccv-four-way-logic}. The incumbent must succeed on the original source, establishing that the behavior is already supported by $S_t$, and the candidate must remain qualified on that source, ruling out the simpler case in which the update has already broken the original task. Previously computed clean results are reused when available; the two \texttt{Evaluate} operations in Algorithm~\ref{alg:ccv} denote logical checks and do not require redundant re-execution.

\paragraph{Candidate-conditioned attack generation.}
After the clean pair qualifies, \texttt{GenerateAttack} receives $e_t$ and the post-complexity candidate change $\Delta_t^{\mathrm{ccv}}$. The generator identifies an assumption, dependency, or decision boundary that the candidate may have introduced, strengthened, or narrowed, and proposes a bounded perturbation designed to stress that property. Generation may return no attack when the candidate change exposes no suitable target or when no valid perturbation can be constructed.

\paragraph{Benchmark-specific attack application.}
The attack specification $a_t$ is converted into an executable instance through \texttt{ApplyAttack}. SpreadsheetBench can perturb workbook cells, formulas, or task instructions subject to validity checks. SearchQA introduces bounded distractor context while retaining the original question and gold answer. LiveMath adds controlled prior scratch-work context while retaining the original problem, answer options, and label. SkillEvolBench preserves the original software-task instruction and appends bounded robustness context targeted at the candidate change. ContinualSkillBench uses both answer-preserving transformations and nearby boundary cases with independently checkable oracles. Although the surface transformations differ, every integration constructs one attacked instance $x_t^{\mathrm{ccv}}$ and holds it fixed for the incumbent--candidate comparison.

\paragraph{Regression criteria.}
The attacked evaluations correspond to the second column of Equation~\ref{eq:ccv-four-way-logic}. A CCV-confirmed regression requires the incumbent to remain sufficiently successful on $x_t^{\mathrm{ccv}}$ while the candidate deteriorates according to the native evaluator. SkillOpt uses a strict incumbent-pass/candidate-fail condition after clean qualification. SkillEvolBench requires the incumbent attacked-case score to be at least $0.8$ and the candidate score to decrease by at least $0.05$ relative to the incumbent; the candidate need not cross a binary failure threshold. ContinualSkillBench requires incumbent success together with either candidate failure or a score decrease greater than $0.1$. These differences affect the numerical realization of \texttt{RegressionCriterion}, while the four-way comparison remains unchanged.

\paragraph{Scope-constrained repair.}
When the paired criterion identifies a regression, CCV uses $\Delta_t^{\mathrm{ccv}}$, $a_t$, and the attacked-case outcomes to localize the candidate content most directly implicated by the failure. SkillOpt and SkillEvolBench restrict repair to the relevant candidate change and related local content, using the local editing mechanism described in Section~\ref{subsec:skillevoreg-complexity}. ContinualSkillBench follows the same locality principle at the level of implicated skill paths and may rewrite the full content of an affected skill. Repair granularity therefore follows the native representation while remaining constrained to the region implicated by the candidate-conditioned attack. Each triggering CCV event performs at most one scope-constrained repair.

\begin{table*}[t]
\centering
\small
\caption{Integration of \method with the three native skill-update interfaces. The regularization mechanisms remain largely unchanged; the main differences concern how the native framework represents, exposes, and writes back a candidate update.}
\label{tab:update-interface-instantiation}
\begin{tabular}{@{}p{0.15\textwidth} p{0.24\textwidth} p{0.24\textwidth} p{0.24\textwidth}@{}}
\toprule
\textbf{Interface} & \textbf{SkillOpt} & \textbf{SkillEvolBench} & \textbf{ContinualSkillBench} \\
\midrule
Skill state & Single evolving instruction document. & Multi-skill library containing localized skill files. & Continually evolving skill library. \\
\addlinespace
Native update & Ranked textual edits generated from the execution trajectory. & Localized skill patch proposed by the native skill author. & Explicit \textsc{Create} or \textsc{Modify} operation. \\
\addlinespace
Candidate representation & Updated candidate document and corresponding textual changes. & Proposed patch over one or more localized skill regions. & Raw post-update skill or library snapshot produced by the native operation. \\
\addlinespace
Delta extraction & Compare the candidate document with the temporary skill view used during update generation. & Recover the changed skill and source patch from the native proposal. & Compare pre-update and raw post-update snapshots and identify affected skill paths. \\
\addlinespace
Dropout reconciliation & Restore masked atoms from the incumbent document and apply only the updater-generated change. & Restore masked library content and reconcile the native patch with the complete incumbent library. & Restore the pre-update library and reconcile the raw \textsc{Create}/\textsc{Modify} result with the restored snapshot. \\
\addlinespace
Local regularization scope & Candidate atoms plus related existing atoms. & Changed skill/source delta plus related local content. & Affected skill paths, with whole-skill rewriting permitted within an implicated path. \\
\addlinespace
Final write-back & Return the regularized candidate to SkillOpt's acceptance and current/best bookkeeping. & Write the finalized patch through the existing library update interface. & Write the final regularized snapshot under the continual framework's apply semantics. \\
\bottomrule
\end{tabular}
\end{table*}

\subsection{Integration with Native Skill Updaters}
\label{app:framework-instantiation-details}

The principal framework-dependent implementation choices occur at the native update boundary. Table~\ref{tab:update-interface-instantiation} summarizes these differences.

The table highlights why the three benchmarks do not require three different formulations of \method. Their native updaters determine how each candidate transition becomes visible and how the final candidate is returned. Once the stage-specific changes are exposed, dropout reconciliation, local structural regularization, and paired counterexample validation retain the same conceptual roles.

\section{Detailed Component and Trajectory Analyses}
\label{app:detailed-analysis}

\subsection{Failure Case: Non-Local Whole-Skill Rewriting}
\label{app:whole-skill-compression-case}

We analyze a SpreadsheetBench failure case to illustrate why \method restricts complexity regularization to candidate-local regions rather than allowing unrestricted whole-skill rewriting. In this case, a validation regression triggered a repair procedure that operated over the complete persistent skill, even though the complexity trigger itself was inactive. The repaired skill was then written back without a post-repair evaluation. We therefore treat this example as a diagnostic case study of non-local rewriting rather than as a controlled comparison between compression strategies.

The original candidate update was comparatively local. Relative to the preceding incumbent, it added four bullets, removed one inherited bullet and one third-level heading, and altered content in only $4$ of the $14$ major \texttt{\#\#} sections. The subsequent repair, however, was permitted to rewrite the complete skill rather than being restricted to this candidate-local change. Although the repair prompt requested localized revision, it exposed the complete skill and imposed no programmatic boundary on the editable region.

\begin{table}[t]
\centering
\small
\caption{Diagnostic failure case illustrating the scope and behavioral consequences of unrestricted whole-skill rewriting. Textual rewrite statistics characterize rewrite scope rather than semantic rule deletion.}
\label{tab:whole-skill-compression-failure}
\begin{tabular}{@{}ll@{}}
\toprule
\textbf{Quantity} & \textbf{Observed change} \\
\midrule
Original candidate scope & $4/14$ major sections affected \\
Whole-skill repair scope & $14/14$ major sections changed \\
Candidate $\rightarrow$ repaired text & $18{,}313 \rightarrow 9{,}598$ characters ($-47.6\%$) \\
Persistent skill length & $2{,}679 \rightarrow 1{,}450$ words ($-45.9\%$) \\
Inherited bullets preserved verbatim & $6/47$ \\
Inherited bullets not preserved verbatim & $41/47$ ($87.2\%$) \\
Test accuracy & $60.5\% \rightarrow 48.5\%$ ($-12.0$ pp) \\
Correct $\rightarrow$ incorrect & 32 tasks \\
Incorrect $\rightarrow$ correct & 8 tasks \\
Post-repair evaluation & None before write-back \\
\bottomrule
\end{tabular}
\end{table}

The rewrite footprint was substantially broader than the source candidate delta. Although the candidate affected only $4/14$ major sections, the whole-skill repair changed content in all $14$ sections, and seven original section titles no longer appeared verbatim. Among the 47 bullets inherited by the candidate from the preceding skill, only six remained textually identical after the repair, while $41/47$ were no longer preserved verbatim. These statistics quantify textual non-locality rather than semantic deletion: they do not imply that 41 distinct rules were removed, since some content may have been merged, relocated, or paraphrased.

The resulting checkpoint also exhibited strongly asymmetric behavioral change. Test accuracy decreased from $60.5\%$ to $48.5\%$: 32 previously solved tasks became failures, while only 8 previously failed tasks became correct. Inspection of the trace indicated that the global rewrite altered established operational guidance rather than merely consolidating candidate-local redundancy. Because the transition also involved an erroneous trigger route and lacked post-repair evaluation, these observations should not be interpreted as a controlled estimate of the causal effect of whole-skill rewriting.

This failure case motivates the separation between the \emph{trigger} for structural reconsideration and the \emph{scope} permitted to change. In \method, complexity measurements can use the broader incumbent and candidate states to determine whether regularization is warranted, but editing is anchored to the candidate-local delta $\mathcal{D}_t$ and a bounded related context $\Omega_t$. Unrelated incumbent content is therefore preserved rather than being exposed to an unconstrained global rewrite. Locality does not require every repair to be small; it requires the permissible edit region to remain tied to the candidate transition being regularized.

\subsection{CCV Regression Evidence and Representative Trace}
\label{app:ccv-details}

\paragraph{Regression evidence.}
Across the evaluated trajectories, CCV identifies 12 distinct regression signals, 9 of which lead to state-changing repairs. Each signal arises from a candidate-conditioned comparison in which the incumbent and candidate remain qualified on the original source, while the candidate deteriorates relative to the incumbent on the same attacked case. These events provide update-level evidence for the type of behavioral regression targeted by CCV. We next examine one representative trace in detail.

\paragraph{Representative CCV trace.}
We use a representative SkillEvolBench dependency-scope trace to illustrate the mechanism. The source task asks the agent to repair a React 18 peer-dependency conflict. The incumbent contains a general instruction to regenerate and validate the repository lockfile. The candidate adds a more specific rule that makes manifest ownership and package/workspace boundaries an explicit decision criterion.

CCV applies the same package-boundary attack to both incumbent and candidate runs. The incumbent receives reward $1.0$, satisfying 7/7 outcome checks and 5/5 process checks. The candidate receives $0.8$: all seven outcome checks remain satisfied, but two process checks fail after it replaces versioned React dependencies with local \texttt{file:} shims. The attacked comparison therefore reveals candidate-specific sensitivity to the package-boundary condition targeted by the candidate-added rule.

CCV rewrites the exact candidate-added atom from a rule that emphasizes the directory owning the manifest to one that ties lockfile regeneration and validation to the dependency scope actually used by the resolver. CCV therefore identifies candidate-specific sensitivity to package-boundary wording and produces a targeted scope-constrained rewrite. A later replay recovers the two affected process checks.

\subsection{Residual Stochasticity and Validation-Based Checkpointing}
\label{app:skillopt-stochasticity}

\paragraph{Shared-prefix continuation.}
Two \method SpreadsheetBench continuations share a byte-identical skill state through step 30, the end of epoch six, and then process the same remaining batch sequence under the same configuration. Steps 29 and 30 are rejected, so the shared step-30 snapshot is also textually identical to the state produced at step 28. The first different skill states appear at step 31, where independently sampled candidate updates are accepted. These are controlled continuations of one learned prefix, not independent end-to-end training seeds.

\begin{table}[t]
\centering
\small
\caption{SpreadsheetBench shared-prefix continuation. E1--E6 are common to
both branches; E7--E8 differ after independent update sampling. Test scores
average two frozen evaluations.}
\label{tab:spreadsheet-branching-full}
\begin{tabular}{clrrr}
\toprule
\textbf{Branch} & \textbf{Checkpoint} & \textbf{Train} & \textbf{Validation} & \textbf{Test} \\
\midrule
\multirow[c]{7}{*}{Shared}
& Initial & 32.50 & 37.00 & 38.75 \\
& E1 & 44.50 & 52.00 & 52.50 \\
& E2 & 48.00 & 55.50 & 56.50 \\
& E3 & 48.00 & 55.50 & 56.50 \\
& E4 & 48.50 & 60.00 & 58.50 \\
& E5 & 48.50 & 60.00 & 58.50 \\
& E6 & 53.50 & 61.00 & 57.25 \\
\midrule
\multirow[c]{2}{*}{Main}
& E7 & 49.50 & \textbf{62.50} & 58.50 \\
& E8 & 50.00 & 61.50 & \textbf{59.25} \\
\midrule
\multirow[c]{2}{*}{Degrading}
& E7 & 46.50 & 51.00 & 52.00 \\
& E8 & 45.50 & 51.50 & 50.00 \\
\bottomrule
\end{tabular}
\end{table}

The validation-selected delivery checkpoint is E7 on the main continuation
and E6 on the degrading continuation. Although the two terminal states diverge substantially ($59.25\%$ versus $50.00\%$ test accuracy), checkpoint selection yields much more comparable delivered performance ($58.50\%$ versus $57.25\%$). This illustrates that regularization and checkpoint selection address distinct sources of risk: \method shapes the repeated update process, whereas validation-based checkpointing protects delivery from residual stochastic trajectory degradation. We retain the latter as an established early-stopping-style safeguard rather than a contribution of \method.

\paragraph{Accepted update patterns.}
After the fork, each branch accepts five of the ten remaining proposals. The main branch accepts steps 31, 32, 33, 35, and 37; the degrading branch accepts 31, 34, 36, 37, and 39. The degrading branch accumulates broader instructions for preserving blank/nonmatching slots, clearing a prefilled destination block, blanking non-terminal rows, and clearing unused spill cells. The main branch is not free of blanking language, but combines it with explicit full-range reconnaissance, blank-versus-zero disambiguation, and conjunctive deletion conditions. The comparison therefore concerns the combination and scope of accepted rules, not a binary distinction between ``blanking'' and ``no blanking.''

\paragraph{Task-level audit.}
The same 200 test tasks are evaluated twice at the shared E6 checkpoint and at the degrading E8 checkpoint. Thirty-six tasks regress, 12 improve, and 152 are unchanged, yielding a net decline of $7.25$ percentage points. Thirty-one of the 36 regressions are cell-level manipulation tasks. Twenty-one of the 36 produce a missing prediction in both E8 repeats, and 30 do so in at least one repeat. The dominant observed failure pattern is therefore missing required cell values rather than uniform degradation across task families.

The accepted clearing and blanking instructions are behaviorally compatible with this pattern, but they are not isolated as the cause. Multiple skill edits and evaluation-time randomness remain confounders. Train and validation performance also decline along the degrading branch. We therefore interpret this controlled fork as stochastic trajectory-level degradation rather than conventional train--test overfitting. The example illustrates a residual source of variation that update-level regularization does not eliminate, and shows how validation-based checkpoint selection can prevent such late-stage degradation from determining the delivered state.

\section{Prompts and Scope Enforcement for LLM-Based Regularization Operations}
\label{app:prompt-templates}

The algorithms in Appendix~\ref{app:regularization-algorithms} define semantic operations such as $\operatorname{SelectOperation}$, $\operatorname{GenerateAttack}$, and $\operatorname{Repair}$ rather than prescribing a single prompt shared across all host frameworks. Their concrete realizations depend on the native skill representation and task-valid perturbation space. This section provides representative prompt excerpts and summarizes the corresponding writable-scope constraints. Native skill-updater prompts are inherited from the host frameworks and are not reproduced here.

For presentation consistency, implementation identifiers in the excerpts are normalized to the terminology used in this paper; the operational instructions and constraints are otherwise preserved. Dynamic runtime fields are shown in braces.

\subsection{Complexity-Aware Local Editing}
\label{app:prompt-complexity}

For SkillOpt and SkillEvolBench, the local editor receives broader skill context for interpretation, but only explicitly identified source-delta regions are writable. The core editing instruction defines four operations:

\begin{quote}
\small\ttfamily\raggedright
Treat the listed source deltas as independent editable regions. Similarity only nominates historical content for semantic comparison; it does not imply that two rules are interchangeable.

For every source delta, choose exactly one operation:

NOOP: preserve the source unchanged when it adds a distinct or complementary capability, or whenever semantic preservation is uncertain.

REWRITE: replace only the source delta when it is overspecific, unclear, or internally redundant, while preserving its supported capability.

MERGE: use only when one coherent rule can preserve the full union of the source delta and linked historical content.

DELETE: remove only a source delta that is unsupported, harmful, or fully subsumed by retained content.

The complete skill is read-only context. Return modifications only for explicitly supplied editable regions; never return a rewritten complete skill.
\end{quote}

When complexity regularization is triggered, the editor additionally receives:

\begin{quote}
\small\ttfamily\raggedright
Merge duplicates and semantic overlaps, remove narrow or unsupported rules, replace similar rule clusters with transferable general rules, and compress repeated explanations or examples. Do not add new rules merely in response to excessive complexity, and preserve distinct useful behavior.
\end{quote}

The corresponding user request separates read-only context from writable regions:

\begin{quote}
\small\ttfamily\raggedright
\#\# Full Current Skill (read-only context)\\
\{current\_skill\}\\[2pt]

\#\# Editable Source-Delta Regions\\
\{editable\_regions\}\\[2pt]

\#\# Regularization Feedback\\
\{regularization\_feedback\}\\[2pt]

Return one decision for every listed source delta and only the replacements required by those decisions.
\end{quote}

SkillEvolBench uses the same four-operation semantics with an additional \texttt{skill\_id}, since edits may belong to different localized skill files. ContinualSkillBench instead follows its native CREATE/MODIFY representation: CREATE operates over explicitly supplied skill paths, whereas MODIFY restricts changes to a set of allowed window identifiers. In all cases, the writable scope is constructed and validated by the framework adapter rather than inferred by the language model.

\subsection{CCV Attack Generation}
\label{app:prompt-ccv-attack}

CCV constructs a candidate-conditioned behavioral probe rather than an arbitrary harder example. Because valid perturbations differ substantially across benchmarks, $\operatorname{GenerateAttack}$ has framework-specific realizations.

SkillOpt uses a two-stage Scout--Worker procedure. The Scout observes the incumbent and candidate skills, their semantic delta, recent trajectories, eligible source tasks, and the verifier contract, and returns a candidate-specific weakness together with a suitable source. A Worker then materializes that weakness using the benchmark's permitted perturbation space. SpreadsheetBench applies bounded operations to a private workbook copy; SearchQA inserts bounded distractor context while preserving the question and gold answer; and LiveMath introduces bounded prior scratch-work context while preserving the original problem, options, and gold label.

SkillEvolBench uses a single bounded probe generator. Its central instruction is:

\begin{quote}
\small\ttfamily\raggedright
Generate one bounded CCV probe for a coding-agent skill update. The original task instruction and verifier are immutable.

Target a specific semantic addition, deletion, or strengthening introduced by the candidate skill. The probe may add only neutral operational context, an answer-free counterfactual condition, or a presentation variation targeted at an incidental assumption introduced by the candidate.

Do not change the requested deliverable, acceptance criteria, repository contents, tests, or environment. Do not reveal a solution, expected patch, hidden test, or answer. Select responsible skill identifiers only from skills changed by the candidate.
\end{quote}

The generator returns the attack type, targeted assumption, predicted failure region, responsible skill identifiers, a goal-preservation justification, and the bounded robustness context appended to the original instruction.

ContinualSkillBench follows the same candidate-conditioned principle but uses domain-specific probe generators for Mathematics, Finance, Law, Healthcare, and Office. These generators target domain-appropriate boundaries or invariance-preserving transformations, with additional oracle or equivalence checks before the probe is used for paired incumbent--candidate evaluation.

\subsection{CCV Repair}
\label{app:prompt-ccv-repair}

Once paired evaluation identifies a candidate-specific regression, the repair stage is instructed to recover a transferable capability rather than encode the attacked example itself. SkillOpt and SkillEvolBench reuse the local editor from Appendix~\ref{app:prompt-complexity} with CCV-specific feedback:

\begin{quote}
\small\ttfamily\raggedright
A CCV regression has been detected. Revise, generalize, or remove the incidental assumption responsible for the regression and restore the exposed transferable capability. Never add a one-off exception for a single counterexample.

Derive exactly one transferable procedural invariant from the CCV counterexample and use it to replace, merge, or generalize the responsible existing rule. Do not copy counterexample-specific field names, file names, schema keys, literal values, concrete formats, or wording; express the invariant at the capability level instead.
\end{quote}

SkillEvolBench additionally provides the measured score decrease, attack type, targeted assumption, and predicted failure region as diagnostic context, while retaining the same region-level editing restriction.

ContinualSkillBench uses a separate repair call at responsibility-path granularity:

\begin{quote}
\small\ttfamily\raggedright
Repair the cause of the detected CCV regression as a transferable invariant. Prefer editing or merging the responsible rule. Modify only the supplied responsibility skills, preserve unrelated semantics, and do not add source answers, task identifiers, or irrelevant probe-specific details.
\end{quote}

Unlike the region-local realizations above, this operation may return the complete text of an authorized responsibility skill. The implementation therefore validates the returned paths and verifies the actual modified-path set before write-back.

\begin{table*}[t]
\centering
\small
\caption{Writable-scope enforcement across host frameworks. The model may inspect broader context, but the adapter determines which regions, windows, or skill paths are eligible for write-back.}
\label{tab:prompt-scope-enforcement}
\begin{tabular}{@{}p{0.19\textwidth}p{0.22\textwidth}p{0.22\textwidth}p{0.24\textwidth}@{}}
\toprule
\textbf{Framework} & \textbf{Readable context} & \textbf{Writable scope} & \textbf{Enforcement} \\
\midrule
SkillOpt &
Complete incumbent skill and local context &
Explicit source-delta regions &
Unknown or duplicate region IDs are rejected; linked historical atoms and merge effects are determined by the runtime. \\
\addlinespace
SkillEvolBench &
Complete same-family skills as read-only context &
Selected regions within implicated skills &
Unknown skill/region pairs and out-of-scope edits are rejected before native commit. \\
\addlinespace
ContinualSkillBench MODIFY &
Source skill and bounded matched context &
Explicit editable windows &
Unknown, stale, overlapping, or out-of-scope windows are rejected before application. \\
\addlinespace
ContinualSkillBench CREATE &
New skill and retrieved historical skills &
Authorized skill paths &
Unauthorized paths are rejected; historical paths can change only under an allowed merge. \\
\addlinespace
ContinualSkillBench CCV repair &
Probe and responsibility-skill contents &
Authorized responsibility paths &
Returned paths and the actual post-repair modified-path set are both checked before write-back. \\
\bottomrule
\end{tabular}
\end{table*}

\subsection{Programmatic Enforcement of Editable Scope}
\label{app:programmatic-scope}

Prompt-level locality is not itself an enforceable editing boundary. As illustrated by the failure case in Appendix~\ref{app:whole-skill-compression-case}, a model can perform a broad rewrite even when instructed to revise locally. In the final implementation, the framework adapter determines the writable scope independently of the model.

The common invariant is therefore not that every repair is atom-local. Rather, the framework adapter determines the writable scope, while the granularity of that scope follows the native representation. SkillOpt and SkillEvolBench use region-level editing, ContinualSkillBench MODIFY uses bounded windows, and ContinualSkillBench CREATE and CCV repair may rewrite complete skills only within explicitly authorized paths.

This separation is deliberate: prompt instructions specify \emph{how} the model should transform authorized content, whereas programmatic scope enforcement determines \emph{what} content is permitted to change.

\section{Extended Related Work}
\label{app:extended-related-work}

\paragraph{Learning from agent experience.} A broad line of research studies how language-model agents can improve through interaction without modifying model parameters. Reflexion stores verbal reflections derived from task feedback in episodic memory and reuses them in subsequent attempts \citep{shinn2023reflexion}, while ExpeL extracts reusable natural-language insights from collections of agent experiences \citep{zhao2024expel}. Agent-Pro iteratively refines a language-level behavioral policy through reflection and search \citep{zhang2024agentpro}. Voyager maintains an expanding library of executable code skills that can be retrieved and composed for later tasks \citep{wang2024voyager}. More recent memory systems explicitly organize reusable experience: Agent Workflow Memory induces recurring workflows from successful trajectories \citep{wang2025awm}, and A-MEM continuously restructures an interconnected memory network as new experiences arrive \citep{xu2025amem}. These methods establish reusable external state as an effective mechanism for non-parametric adaptation; our work focuses on regularizing repeated revision of such procedural state.

\paragraph{Skill generation, optimization, and maintenance.} SkillOpt formulates a natural-language skill as trainable external state and optimizes it through bounded textual edits, validation-gated acceptance, and checkpoint selection \citep{yang2026skillopt}. Trace2Skill aggregates trajectory-local lessons before hierarchically consolidating them into transferable skills \citep{ni2026trace2skill}, while CoEvoSkills jointly evolves structured skill packages and a surrogate verifier \citep{zhang2026coevoskills}. SkillEvolBench evaluates transfer from episodic experience to procedural skills under context shifts, adversarial shortcuts, and skill composition \citep{lei2026skillevolbench}; ContinualSkillBench studies skill creation and modification across sequential tasks and highlights the difficulty of consolidating experience into compact reusable skills \citep{guan2026continualskillbench}. Closest to our reliability motivation, GSE uses a global skill-relation graph, cross-task consolidation, and replay-driven verification \citep{yang2026gse}; SkillCommit validates broader abstractions before committing them \citep{he2026skillcommit}; and SkillAdam uses optimization history and adaptive edit budgets to stabilize iterative skill optimization \citep{li2026skilladam}. These methods provide safeguards within particular evolution algorithms, whereas \method asks whether complementary anti-overfitting principles can transfer across different update semantics.

\paragraph{Optimization of prompts and external language-model state.} ProTeGi uses textual feedback analogous to gradients together with beam search to improve prompts \citep{pryzant2023protegi}; OPRO treats an LLM itself as an optimizer over natural-language solutions \citep{yang2024opro}; EvoPrompt and Promptbreeder use evolutionary search over prompts \citep{guo2024evoprompt,fernando2024promptbreeder}; and PromptAgent formulates prompt optimization as strategic search \citep{wang2024promptagent}. DSPy extends optimization to modular language-model programs \citep{khattab2024dspy}, while TextGrad propagates language-model feedback through compound AI systems in analogy to automatic differentiation \citep{yuksekgonul2025textgrad}. These works primarily address how to search for higher-performing external state. \method instead regularizes the transitions proposed by an existing updater.

\paragraph{Regularization and counterexample-based validation.} Classical learning combines multiple mechanisms to control overfitting: dropout reduces feature co-adaptation \citep{srivastava2014dropout}, weight decay constrains unnecessary capacity \citep{krogh1991weightdecay}, and data augmentation or adversarial training exposes models to informative perturbations \citep{goodfellow2015adversarial,zhang2018mixup}. Validation-based early stopping provides an additional safeguard against late-stage deterioration \citep{prechelt1998earlystopping}. \method transfers the principles behind these techniques to discrete skill evolution rather than reproducing their parametric implementations. CCV is also related to counterexample-guided synthesis, differential testing, and program repair, where carefully chosen failures reveal weaknesses in candidate programs. Counterexample-guided program repair, for example, combines failing examples with fault localization to direct corrections \citep{orvalho2025counterexample}. In our setting, the validated object is a skill transition: incumbent and candidate are first qualified on the original source and then compared on the same candidate-conditioned attacked case, turning the counterexample into an update-level behavioral check.

\end{document}